%% file: main.tex
\documentclass[11pt]{article}

\usepackage[preprint]{acl}

\usepackage{times}
\usepackage{latexsym}
\usepackage{enumitem}
\usepackage{booktabs}
\usepackage{amssymb}
\usepackage{array}
\usepackage{tabularx}
\usepackage{listings}
\usepackage{graphicx}
\usepackage{makecell}
\usepackage[table]{xcolor}

\lstdefinestyle{prompt}{
  basicstyle=\ttfamily\scriptsize,
  breaklines=true,
  columns=fullflexible,
  frame=single,
}

\usepackage[T1]{fontenc}

\usepackage[utf8]{inputenc}

\usepackage{microtype}

\usepackage{inconsolata}

\usepackage{graphicx}
\newcommand{\revise}[1]{#1}
\newif\ifcameraready
\camerareadytrue

\ifcameraready
  \newcommand{\siteA}{\revise{Baltanas}}
  \newcommand{\siteB}{\revise{Delos}}
  \newcommand{\siteC}{\revise{Lucretili}}
  \newcommand{\siteD}{\revise{Ranverso}}
  \newcommand{\siteE}{\revise{Schenkenberg}}
\else
  \newcommand{\siteA}{site~A}
  \newcommand{\siteB}{site~B}
  \newcommand{\siteC}{site~C}
  \newcommand{\siteD}{site~D}
  \newcommand{\siteE}{site~E}
\fi
\title{HeriKGQA: Grounded Knowledge-Graph Question Answering over Cultural Heritage Sensor Data}

\author{
 \textbf{Lingxiao Kong \textsuperscript{1,2}},
 \textbf{Vasileios Arampatzakis \textsuperscript{3}},
 \textbf{Vasileios Sevetlidis \textsuperscript{3}},\\
 \textbf{George Pavlidis \textsuperscript{3}},
 \textbf{Zeyd Boukhers \textsuperscript{1,4}}
\\
 \textsuperscript{1} Fraunhofer Institute for Applied Information Technology FIT,
 \textsuperscript{2} University of Cologne,\\
 \textsuperscript{3} Athena Research and Innovation Centre,
 \textsuperscript{4} University Hospital of Cologne
\\
 \small{
   \textbf{Correspondence:} \href{mailto:lingxiao.kong@fit.fraunhofer.de}{lingxiao.kong@fit.fraunhofer.de}, \href{mailto:zeyd.boukhers@fit.fraunhofer.de}{zeyd.boukhers@fit.fraunhofer.de}
 }
}

\begin{document}
\maketitle
\begin{abstract}
Cultural heritage sites are increasingly monitored by sensor networks, but querying their data requires database expertise. Large Language Models (LLMs) offer a natural-language interface, yet over raw sensor records they lose track of which sensor, site, and time each value belongs to, and \textit{silently miscompute}. We present \textbf{HeriKGQA}, a knowledge-graph question-answering \revise{method} that confines the LLM to interpreting questions and phrasing answers, leaving computation to SPARQL and deterministic code. Questions are parsed into intents constrained by a heritage ontology aligning W3C SOSA/SSN with CIDOC-CRM, compiled into SPARQL, and answered from query results and temporal, spatial, and anomaly facts derived by a grounding layer, with error correction guarding each LLM step. On 500 questions over real data from five heritage sites, HeriKGQA reaches 99.0\% accuracy, 18.2 points above the strongest of three baselines using the same LLM and receiving the same anomaly rules and answer conventions, with comparable answer quality while running almost four times faster. Without grounding, accuracy drops by 12.2 points and on anomaly questions from 99\% to 66\%. We report lessons from development and live deployment.
\end{abstract}

\input{sections/1_introduction}
\input{sections/2_problem}
\input{sections/3_method}
\input{sections/4_experiments}
\input{sections/5_deployment}
\input{sections/6_related_work}
\input{sections/7_conclusion}

\section*{Acknowledgment}
This work has been supported by the ARGUS EU project (Grant Agreement No. 101132308), funded by the European Union. Views and opinions expressed are however those of the author(s) only and do not necessarily reflect those of the European Union or of the European Research Executive Agency (REA). Neither the European Union nor the granting authority can be held responsible for them. \revise{We thank Ziyue Chen and Ruiyu Ren (University of Cologne) for the human evaluation.}


\bibliography{custom}

\appendix
\input{sections/appendix}

\end{document}

%% file: sections/1_introduction.tex
\section{Introduction}

\begin{figure}[t]
   \begin{center}
\includegraphics[width=1.0\linewidth]{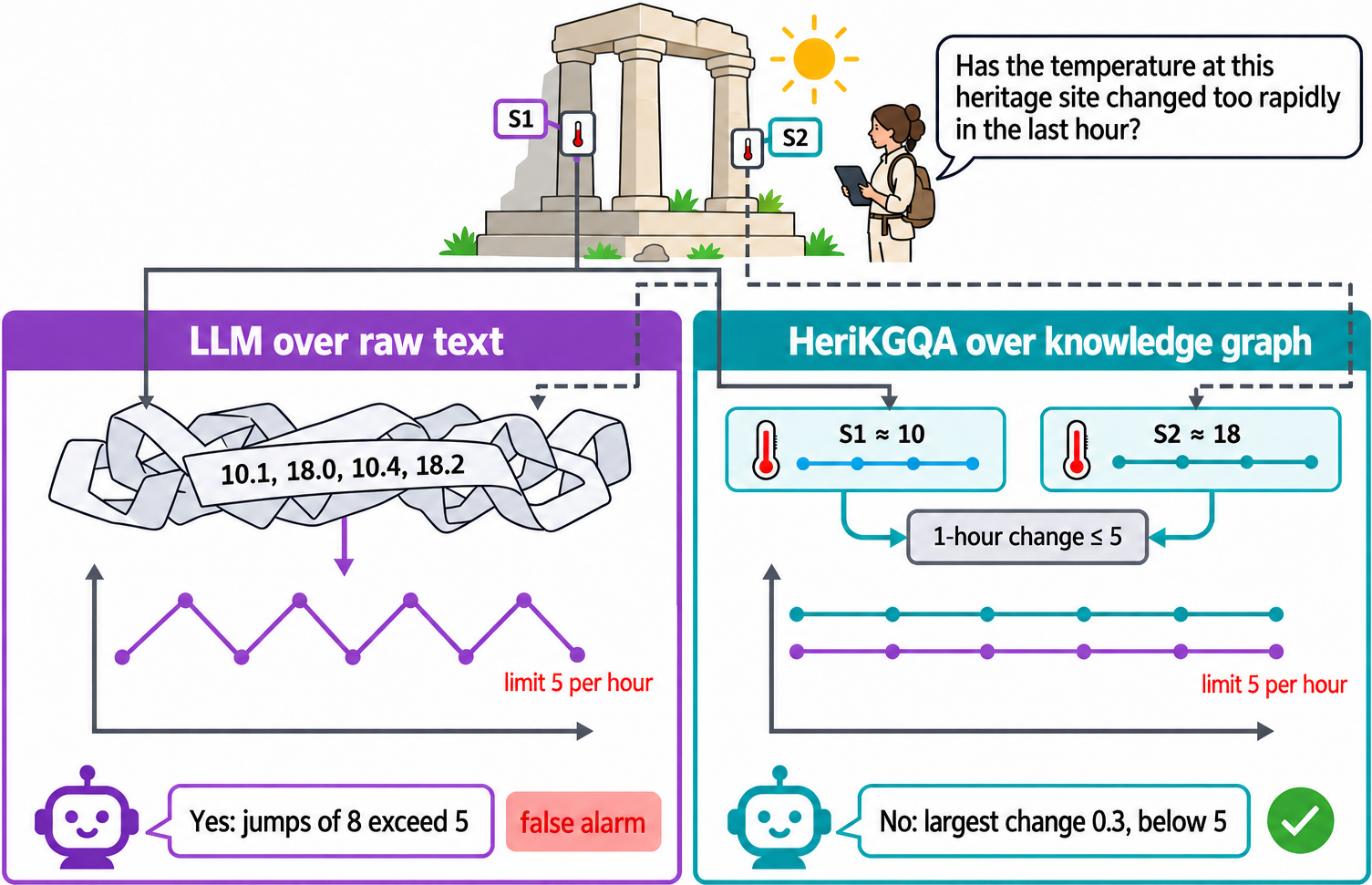}
   \end{center}
   \vspace{-0.5em}
   \caption{\textbf{Example challenge of QA over heritage data.} Over raw text, the LLM mixes two sensors' readings and raises a false alarm (left); HeriKGQA evaluates sensors separately over the knowledge graph (right).}
   \vspace{-0.5em}
\label{fig:task}
\end{figure}

Cultural heritage sites are increasingly monitored by sensor networks that record temperature, humidity, and vibration~\cite{bhatta2024use, rossi2023structural}. Conservators ask questions such as \texttt{``Has the temperature changed too rapidly in the last hour?''} (Figure~\ref{fig:task}), which are easy to state but require filtering, aggregating, and checking thousands of raw readings against expert-defined thresholds. Large Language Models (LLMs) offer a natural-language interface, but over readings serialised as text, they lose track of which sensor, site, and time each value belongs to and miscompute aggregates~\cite{liu2024rethinking}. These errors are \emph{silent}: a fluent answer hides a miscomputed number and can thus mislead conservators deciding whether a monument is at risk.

Retrieval-Augmented Generation (RAG) only selects which readings enter the prompt, leaving the LLM to compute over them. Table-oriented variants like TableRAG~\cite{chen2024tablerag} execute LLM-written pandas code over retrieved cells, and LLM-based Knowledge Graph Question Answering (KGQA) methods such as Interactive-KBQA execute LLM-written SPARQL, including aggregate queries~\cite{xiong2024interactive}. Their answers are thus only as correct as the LLM-written code or queries. Designed for general-domain tables and graphs~\cite{pan2024unifying, sun2024think}, these methods do not address spatial, temporal, and anomaly-related questions over heritage sensor data, whose answers need exact computation.

We present \textbf{HeriKGQA}\footnote{The code, dataset snapshot, and QA benchmark are publicly available at \url{https://github.com/engineerkong/ARGUS-KGQA}. HeriKGQA \revise{powers the chatbot} of the ARGUS platform (\url{https://argus.athenarc.gr/}).}, a KGQA \revise{method} built on the principle that \emph{the LLM reasons over data but never computes over it}. Rather than letting the LLM write free-form code or queries, HeriKGQA restricts the LLM to a constrained intent, compiles it deterministically into SPARQL, and computes spatial, temporal, and rule-based results with pre-implemented, domain-specific functions in a grounding layer; the LLM then verbalises the answer from these results. Its knowledge graph maps the sensor database onto a heritage ontology that makes each reading's relations explicit, and the intent enforces their use. Our contributions are:

\begin{itemize}[leftmargin=*]
\item We realise this principle in HeriKGQA through ontology-constrained intent parsing, deterministic SPARQL compilation, a grounding layer, and LLM error correction (Section~\ref{sec:method}).
\item We construct a 500-question benchmark over real data from five heritage sites, with validated answers, on which HeriKGQA outperforms three baselines and two ablations (Section~\ref{sec:experiments}).
\item We deploy HeriKGQA \revise{in the ARGUS chatbot} for live heritage monitoring, evaluate it on real user questions, and report lessons (Section~\ref{sec:deployment}).
\end{itemize}

%% file: sections/2_problem.tex
\section{Problem Setting}
\label{sec:problem}

We consider five heritage sites, \siteA{}, \siteB{}, \siteC{}, \siteD{}, and \siteE{} (Table~\ref{tab:data}(a)), monitored by 71 sensor nodes. Readings arrive as raw, mostly unitless device channels whose names are reused across sensors (at \siteA{} alone, 19 sensors report \texttt{temperature}), so a reading is interpretable only together with its site, node, sensor, timestamp, and position. Conservation experts also defined 37 anomaly rules, each comparing a computation over a quantity against a site-specific threshold; some refer to quantities, such as crack width or tilt, that no deployed sensor measures.

Conservators and site managers ask spatial, temporal, and anomaly-related questions, alone or as follow-ups to earlier turns. Formally, given a knowledge graph $\mathcal{G}$ of sites, nodes, sensors, observations, and anomaly rules, a question $q$, and an optional history $H$, the task is to produce an answer $a$ whose requested quantitative claim (e.g., a count, average, extremum, range, distance, duration, or threshold verdict) is \emph{correct with respect to $\mathcal{G}$}. When $\mathcal{G}$ lacks the data needed, $a$ must say so instead of producing a number, and when an applicable rule cannot be evaluated and no rule fires, an anomaly answer must be \emph{Cannot be determined}.

A tempting baseline is to retrieve relevant readings and let an LLM answer from them directly, which fails silently in two ways. First, \textit{the readings are serialised as plain text}, where the relations linking each value to its sensor, site, time, and rule are no longer explicit: two temperature sensors at \siteC{} reading steadily near 10 and 18, once interleaved, appear to jump by 8 at every step, triggering a false alarm against the 1-hour threshold of 5 (Figure~\ref{fig:task}). Second, \textit{the LLM computes over raw values itself}, which it does unreliably~\cite{liu2024rethinking}: \siteA{} alone produces about 7{,}760 readings per day, so the model either answers from a truncated subset or must aggregate thousands of values; \texttt{``the last hour''} must be anchored at the latest observation, not the current time, as data arrive with delays and gaps; and spatial questions, such as finding the closest pair, require 171 haversine distances among the 19 nodes at \siteB{}.

%% file: sections/3_method.tex
\section{Method}
\label{sec:method}

\begin{figure*}[t]
   \begin{center}
\includegraphics[width=1.0\linewidth]{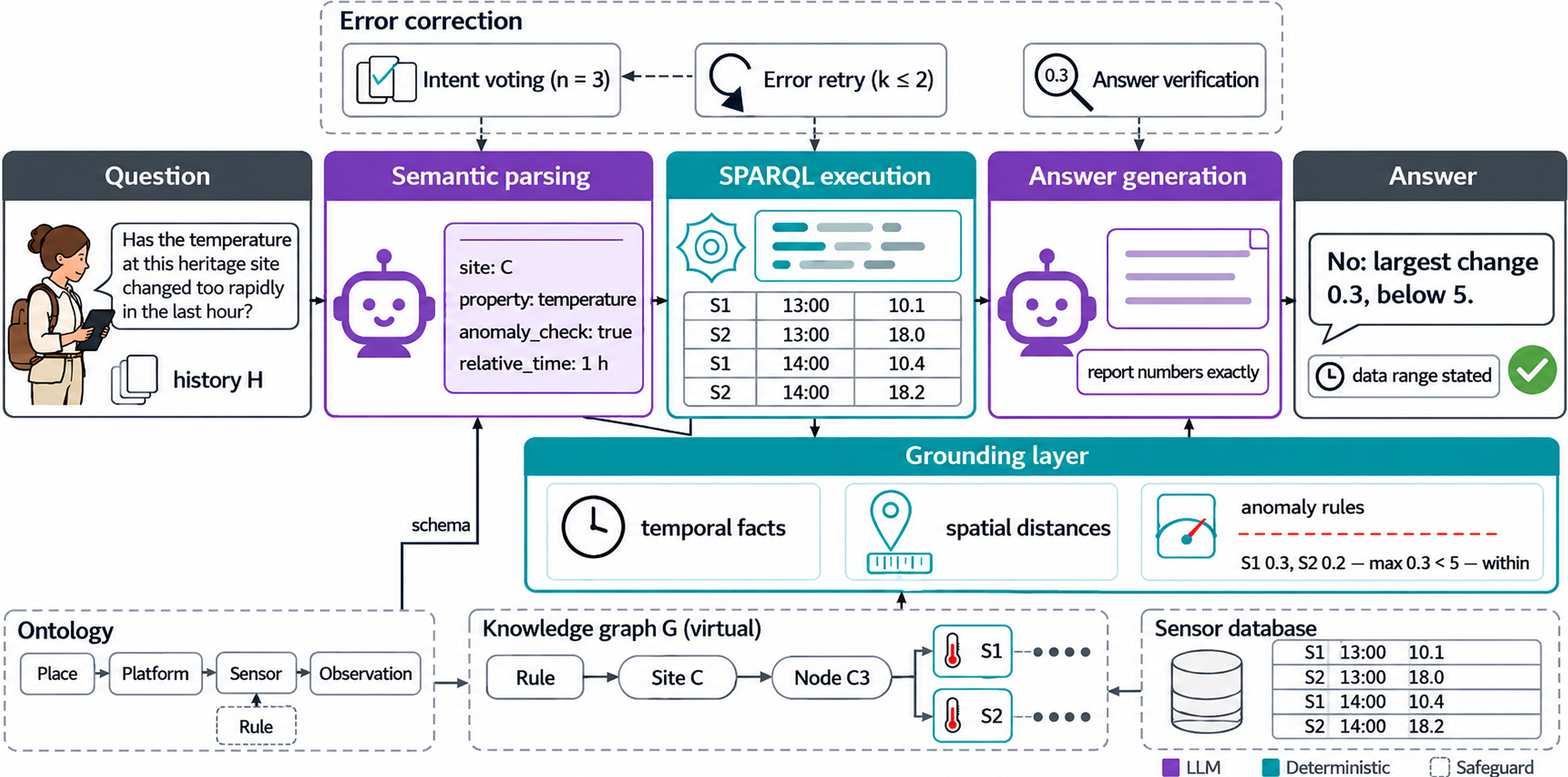}
   \end{center}
   \vspace{-0.5em}
   \caption{\textbf{Overview of HeriKGQA.} \textit{Bottom}: sensor readings are mapped through a heritage ontology, extended with anomaly rules, into a virtual knowledge graph $\mathcal{G}$. \textit{Middle}: the LLM (purple) only parses the question and verbalises the answer, while SPARQL and the grounding layer (teal) compute every number from $\mathcal{G}$. \textit{Top}: voting over intents, retrying failed queries, and verifying answer numbers guard the LLM steps.}
   \vspace{-0.5em}
\label{fig:method}
\end{figure*}

HeriKGQA (Figure~\ref{fig:method}) answers questions over a knowledge graph (Section~\ref{sec:method-kg}) in three steps (Section~\ref{sec:method-pipeline}), supported by a grounding layer and guarded by error correction (Section~\ref{sec:method-layer}).

\subsection{Sensor Knowledge Graph}
\label{sec:method-kg}

\revise{We build the sensor knowledge graph of HeriKGQA on PANOPTES}~\cite{pavlidis2025panoptes}, a digital twin ontology for cultural heritage assets that aligns SOSA/SSN~\cite{janowicz2019sosa, haller2018modular} with CIDOC-CRM~\cite{doerr2003cidoc}. We map each sensor reading to a \texttt{sosa:ObservationCollection} with one \texttt{sosa:Observation} per reported quantity, linked to its sensor, node, and site, and each anomaly rule to a PANOPTES \texttt{Policy} linked to its site (Table~\ref{tab:data}(b)). Every value thus keeps the relations that plain-text rows lose (Section~\ref{sec:problem}): its sensor, so that anomaly checks run per sensor and observed property even when channel names are shared; its node's WGS84 coordinates and altitude, for distances; its timestamp, for time windows; and its site's rules, for site-specific thresholds. Ontop~\citep{xiao2020virtual} serves the graph virtually by translating each SPARQL query into SQL, so new readings are queryable immediately.

\subsection{Three-Step Pipeline}
\label{sec:method-pipeline}

\textbf{Semantic parsing.} An LLM maps the question to a JSON intent: a target class with filters, return properties, joins, and an optional aggregation, ordering, and limit. Relative expressions such as ``the last hour'' become a \texttt{relative\_time} amount and unit, and questions about information the graph does not model are marked \texttt{unanswerable} and declined without a query. The parser is \emph{schema-constrained}: it sees only the instantiated ontology classes and all property, site, and node labels in $\mathcal{G}$, so its filters use the graph's own names. Its prompt also states the reference time (the latest observation in $\mathcal{G}$) and keeps \emph{computation out of the parser}: statistics must be expressed as aggregations, and distance questions return only coordinates. \emph{Anomaly questions} are marked with an \texttt{anomaly\_check} flag and an optional \texttt{as\_of} time, and target the anomaly rules of the named site for the named quantity; for each retrieved rule, a second query fetches that quantity's observations within the rule's window, which the grounding layer evaluates per sensor. \emph{Follow-up questions} are resolved into standalone intents using the last $h$ question--intent pairs (Appendix~\ref{app:implementation}).

\textbf{SPARQL execution.} The intent is validated and \emph{compiled deterministically} into SPARQL, with joins as triple patterns, filters as \texttt{FILTER} expressions, and aggregations as \texttt{GROUP BY} clauses, so all aggregates are computed by the query engine. Unlike free-form SPARQL written by an LLM, an intent can be checked against the schema and voted over, and its compiled queries follow fixed patterns: relative windows end at the latest observation rather than the wall clock, and each aggregate returns its number of values, so an aggregate over no data is treated as empty rather than reported as 0. Since truncating \emph{rows passed to the LLM} would cause silent errors, more than 1,000 such rows prompt the user for an aggregate or a narrower question.

\textbf{Answer generation.} A second LLM call verbalises the answer from the question, the query results, and the facts and verdicts computed by the grounding layer (Section~\ref{sec:method-layer}). Its prompt requires reporting aggregates and verdicts \emph{exactly as given}, never recomputing them from the retrieved rows or presenting a limited result as complete, and stating when no data or rule was found. Anomaly answers begin with \texttt{Yes}, \texttt{No}, or \texttt{Cannot be determined}, and every answer ends with a deterministically appended note on the time range of the data used.

\subsection{Grounding and Error Correction}
\label{sec:method-layer}

\textbf{Grounding layer.} SPARQL covers statistics, but spatial, temporal, and anomaly questions need derived facts, such as distances or rule verdicts, that the answer LLM would otherwise compute. This layer therefore derives them from all returned rows, each only when the rows contain its inputs: \emph{temporal} facts (oldest and newest timestamp, time span, and age of each reading), \emph{spatial} facts (centroid, bounding box, altitude range, and nearest and farthest node pairs by haversine distance), and \emph{numeric} summaries (count, minimum, maximum, and mean) of the listed distinct values per property, never merging quantities. For \emph{anomaly questions}, each retrieved rule is evaluated by type (a threshold on the latest value, a windowed sum, mean, or range, a z-score, or a custom metric such as the rate of change) per sensor and observed property, with windows ending at the rule's latest matching reading (at or before \texttt{as\_of}, if given). Per-sensor results are combined by maximum or minimum as the rule requires and compared with the site-specific threshold, yielding \texttt{exceeds} or \texttt{within} with supporting values; rules lacking an implementation or matching data are flagged without a verdict. Verdicts and references follow the implemented rules, some approximating the expert definitions (Appendix~\ref{app:implementation}), rather than an independent risk assessment.

\textbf{Error correction.} The two LLM steps can still fail: a misparsed intent yields a wrong or empty query, which surfaces as missing data, and the answer LLM may alter a number while phrasing it. Three mechanisms guard against this. \emph{Error-driven retry}: if an intent is invalid or its query fails or returns nothing, the error, query, and intent are fed back to the parser for at most $k$ retries. \emph{Intent voting}: the parser samples $n$ intents in parallel, the greedy parse and $n-1$ at temperature 0.7, and keeps the most frequent, breaking ties in favour of the greedy parse, a form of self-consistency~\citep{wang2022self}. \emph{Answer verification:} every number in the answer is checked against the rows, facts, verdicts, and question (for its occurrence, not for its association with a sensor, site, or time), and an anomaly answer must lead with the computed verdict; a failing answer is regenerated once and otherwise replaced by a deterministic template answer. Deterministic compilation ensures that the query executes the parsed intent, not that the intent matches the question, so parsing errors remain the main source of errors (Appendix~\ref{app:analysis}).

%% file: sections/4_experiments.tex
\section{Offline Experiments}
\label{sec:experiments}

\begin{table*}[t]
\centering
\resizebox{\textwidth}{!}{%
\begin{tabular}{@{}l>{\columncolor{gray!12}}c@{\hspace{14pt}}ccccccccc@{}}
\toprule
& \multicolumn{6}{c}{Accuracy (\%)} & \multicolumn{2}{c}{Quality$^\dagger$} & \multicolumn{2}{c}{Latency (s)} \\
\cmidrule(lr){2-7} \cmidrule(lr){8-9} \cmidrule(lr){10-11}
System & Mean & Basic & Spatial & Temporal & Anomaly & Follow & Mean $\uparrow$ & $<$4 (\%) $\downarrow$ & Median & p90 \\
\midrule
TextRAG               & 37.6 & 17 & 63 & 26 & 63 & 19 & 4.72 & 8.4 & \textbf{3.9} & \textbf{12.4} \\
TableRAG              & 48.2 & 76 & 22 & 57 & 26 & 60 & 4.39 & 32.2 & 8.9 & 17.1 \\
Interactive-KBQA      & 80.8 & 95 & 54 & 91 & 79 & 85 & \textbf{4.85} & 6.2 & 46.7 & 130.9 \\
\midrule
HeriKGQA   & \textbf{99.0} & \textbf{99} & \textbf{100} & \textbf{98} & \textbf{99} & \textbf{99} & 4.79 & \textbf{3.2} & 12.0 & 22.9 \\
\quad w/o grounding  & 86.8 & \textbf{99} & 83 & \textbf{98} & 66 & 88 & 4.78 & 9.8 & 11.4 & 24.4 \\
\quad w/o error correction  & 97.6 & 97 & \textbf{100} & \textbf{98} & \textbf{99} & 94 & 4.82 & \textbf{3.2} & 8.1 & 18.4 \\
\bottomrule
\end{tabular}%
}
\caption{Answer accuracy, quality, and latency. Accuracy \emph{Mean} (shaded) averages the five question types. Ablations remove the grounding layer or the error-correction mechanisms. Quality is rated by an LLM judge (Appendix~\ref{app:quality}): \emph{Mean} averages the Readability, Conciseness, and Comprehension scores (1--5), and \emph{$<$4} is the share with any category scored below 4. Best values in bold. $^\dagger$Human evaluators agree with the judge on 79\% of its non-tie verdicts and prefer HeriKGQA (66 vs.\ 28 of 100 questions sampled across judge verdicts; Table~\ref{tab:human-eval}).}
\label{tab:main-results}
\end{table*}

\textbf{Data and benchmark.} Experiments run on a fixed snapshot of the sensor database covering the five sites, 71 nodes, and 37 anomaly rules of Section~\ref{sec:problem}. The snapshot spans 13 May--12 August 2026, contains 5{,}227{,}617 observations, and is served by Ontop as a virtual knowledge graph over a read-only database copy. We evaluate on 500 questions, 100 each of five types: basic, spatial, temporal, anomaly-related, and follow-up (Appendix~\ref{app:benchmark}).

\textbf{Baselines and ablations.} We compare HeriKGQA against three baselines: (1)~TextRAG, which serialises retrieved readings as plain-text rows; (2)~TableRAG~\citep{chen2024tablerag}, which retrieves columns and cells from the snapshot flattened into one table and answers by executing LLM-written pandas code; and (3)~Interactive-KBQA~\citep{xiong2024interactive}, which has the LLM explore the knowledge graph over multiple turns through SPARQL-based tools. We also ablate the grounding layer (\textit{w/o grounding}), leaving the answer LLM to compute the facts and rule verdicts from the retrieved rows, and error correction (\textit{w/o error correction}: one greedy parse, without retries or answer verification). All six systems use the same LLM (\texttt{gpt-oss-120b}) via \revise{an institutional LLM endpoint}. They differ in the setup around the LLM: schema exposure, demonstrations, tools, and result handling (Table~\ref{tab:systems}). All systems receive the same answer conventions (lead with the verdict; say when the data cannot answer the question) and the rule definitions, which TableRAG finds as rows of its table. HeriKGQA anchors relative time windows deterministically at the latest observation; the baselines must infer the anchor: TextRAG is told the period the data cover, TableRAG’s table caption gives its date range, and Interactive-KBQA can query the latest timestamp (Appendix~\ref{app:implementation}).

\textbf{Evaluation metrics.} \emph{Answer accuracy} counts an answer as correct only if the requested quantitative claim (Section~\ref{sec:problem}) matches a manually validated reference computed independently from $\mathcal{G}$, anomaly answers give the correct verdict, and unanswerable questions are declined. Verdicts count wherever they appear in the answer and are extracted by an LLM classifier. \emph{Answer quality} is rated by an LLM judge on readability, conciseness, and comprehension following FLASK~\citep{ye2024flask}, and complemented by a human pairwise evaluation against Interactive-KBQA (Appendix~\ref{app:quality}). \emph{Latency} is the end-to-end time per question (median, p90).

\textbf{Main results.} As shown in Table~\ref{tab:main-results}, HeriKGQA answers 495 of the 500 questions correctly (99.0\%), 18.2 points above the strongest baseline, Interactive-KBQA (80.8\%), and reaches at least 98\% on every question type, whereas each baseline falls below 60\% on at least one. The baselines fail wherever computation is left to the LLM, either directly (TextRAG) or through code and queries it writes (TableRAG, Interactive-KBQA). TextRAG reaches only 17\% on basic questions, as it must derive statistics from limited retrieved text. TableRAG recovers them through pandas code (76\%), but its code rarely reproduces distances or rule evaluations (22\% spatial, 26\% anomaly). Interactive-KBQA, despite querying the same graph, performs well where SPARQL aggregates suffice (95\% basic, 91\% temporal) but drops on spatial (54\%) and anomaly (79\%) questions, which require distances and per-sensor rule evaluations that it must express in its own SPARQL or estimate from retrieved values. Appendix~\ref{app:robustness} tests the \textit{generality} of our design on MultiTQ~\cite{chen2023multi}, where HeriKGQA is competitive with Interactive-KBQA (75.4\% vs.\ 73.4\%), and its \textit{robustness} to a smaller backbone (\texttt{qwen3-8-27b}) with unchanged accuracy.

\textbf{Ablations.} Without the grounding layer, the answer-generation LLM still receives the rule definitions, computation conventions, and readings each rule applies to, but must compute the facts and verdicts itself. Accuracy then falls by 12.2 points to 86.8\%: basic and temporal questions are barely affected (99\% and 98\%), as their numbers still come from SPARQL aggregates, but spatial and anomaly accuracy drop to 83\% and 66\%. 29 of the 34 anomaly errors are declines, as the raw readings the LLM would need exceed the 1,000-row limit; on the 41 verdict questions whose readings fit, the LLM decides 36 correctly, against 41 with grounding (Appendix~\ref{app:analysis}). The drop thus reflects both the grounding layer's computation and its answer check, which cannot run without the computed facts. Compared with Interactive-KBQA, which queries the same graph with the same LLM, the compiled intent (w/o grounding) raises spatial accuracy from 54\% to 83\% and overall accuracy from 80.8\% to 86.8\%. Removing error correction costs 1.4 points (97.6\%), mostly on basic and follow-up questions (97\% and 94\%): 7 of its 12 errors follow an empty query result, and with error correction 6 of these 7 are answered correctly through an error-driven retry or intent voting.

\textbf{Answer quality and latency.} Confining the LLM to verbalisation does not reduce answer quality: HeriKGQA scores 4.79 out of 5, on par with Interactive-KBQA (4.85), with the fewest answers rated below 4 in any category (3.2\%). TableRAG scores lowest (4.39), as its answers are concise but often incomplete (Comprehension 4.00; Table~\ref{tab:quality-by-type}), and removing the grounding layer triples the share of low-rated answers (from 3.2\% to 9.8\%). HeriKGQA answers in a median of 12.0 s (p90 22.9 s), 3.9 times faster than Interactive-KBQA (46.7 s); the faster TextRAG and TableRAG (median 3.9 s and 8.9 s) stay below 50\% accuracy.

%% file: sections/5_deployment.tex
\section{Online Deployment}
\label{sec:deployment}
\begin{figure}[t]
   \begin{center}
\includegraphics[width=1.0\linewidth]{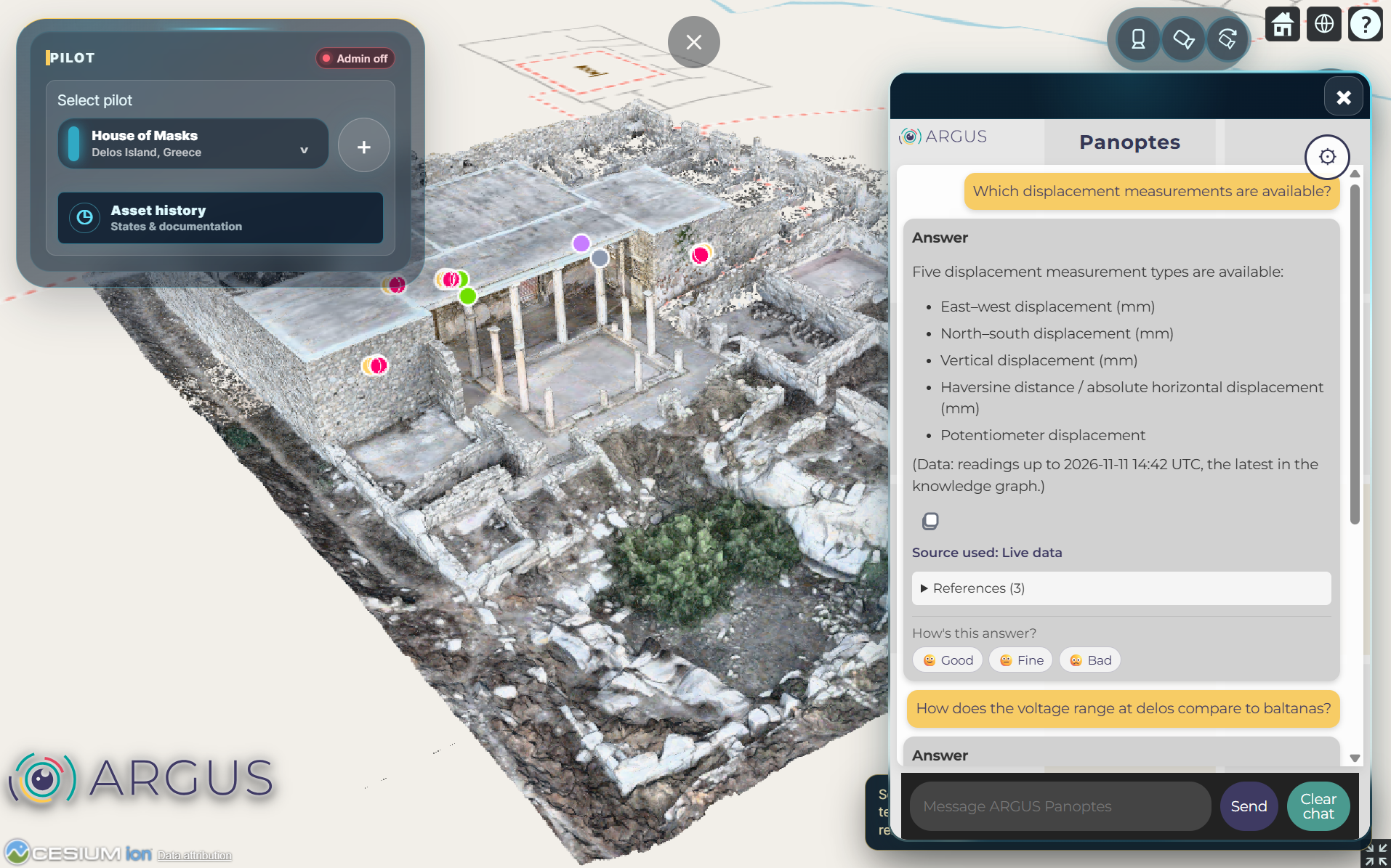}
   \end{center}
   \vspace{-0.5em}
   \caption{\textbf{\revise{HeriKGQA deployed in the chatbot of} the ARGUS platform}. Users select a pilot site (top left; here the House of Masks, Delos), shown as a 3D model with its sensor nodes, and ask questions in the chat panel (right). Each answer states the time of the latest reading used and its data source, with an expandable \emph{References} section holding the parsed intent, SPARQL query, and retrieved data behind it, and a feedback prompt.}
   \vspace{-0.5em}
\label{fig:deployment}
\end{figure}

\textbf{Deployment setup.} On the ARGUS project's infrastructure, HeriKGQA \revise{answers sensor-data questions in the platform's chatbot,} alongside a document-based RAG service for general questions about the sites; its data layer runs as four Docker services: a PostgreSQL database storing the sensor data, an ingestion service that polls live readings from the site gateways, another that loads the context data (sites, nodes, sensors, and rules), and an Ontop~5.5 server~\citep{xiao2020virtual} that exposes the database as a virtual SPARQL endpoint through an OBDA mapping onto PANOPTES. HeriKGQA queries this endpoint, so newly ingested readings are queryable immediately (Section~\ref{sec:method-kg}), and accesses the LLM through the same institutional endpoint as in Section~\ref{sec:experiments}. \revise{Each answer includes an expandable \emph{References} section that shows its parsed intent, SPARQL query, and retrieved data} (Figure~\ref{fig:deployment}), so users can inspect how it was computed.

\textbf{Lessons from development.} Three issues on real sensor data shaped the design. \textit{Vocabulary mismatch:} users name quantities differently from device channels (e.g., ``soil temperature'' for \texttt{soil\_temperature}), which motivated giving the parser all entity and property labels in $\mathcal{G}$. \textit{Label resolution:} resolving result URIs to labels one at a time dominated query time, and batching them into \texttt{VALUES} queries of 50 URIs halves retrieval time for a 200-row result. \textit{LLM capacity:} when we first let the LLM compute averages from the retrieved rows, its results on 12 test questions were off by 4\% on average and up to 21\%, which motivated computing requested statistics in SPARQL.

\textbf{Evaluation on real user questions.} From 21 August to 14 October 2026, the deployment logged 618 queries from project partners and site managers; 239 were routed to HeriKGQA, the others to the document service or general replies. After removing duplicates, misrouted, and overly vague questions, we kept 80 questions about the sensor data and replayed them retrospectively: all systems were re-run over the frozen snapshot (the two follow-ups with their preceding turns) and scored against references computed under benchmark conventions. This measures replay accuracy, not that of the original live answers. HeriKGQA answers 78.8\% (63/80) correctly, ahead of its ablations without grounding (73.8\%) and error correction (70.0\%) and of Interactive-KBQA (57.5\%). Error correction contributes more here than grounding (7 vs.\ 4 questions), as free phrasing more often uses labels not in the graph. HeriKGQA and Interactive-KBQA both correctly decline the 11 questions about data or rules the snapshot lacks, so the gap comes from the remaining 69 questions, which need a substantive answer (52 vs.\ 35 correct); HeriKGQA's 17 errors mostly involve places not modelled as sites in the knowledge graph, results above the 1,000-row cap, and long answer lists.

%% file: sections/6_related_work.tex
\section{Related Work}
\label{sec:related}

\textbf{Cultural heritage monitoring.} Ontologies combining SOSA/SSN with CIDOC-CRM link sensor data to heritage documentation, from sensor-enabled digital twins~\cite{niccolucci2024digital} to \revise{our previous ontology} PANOPTES~\cite{pavlidis2025panoptes}. SigSPARQL~\cite{schwarzinger2025sigsparql} adds signal operators to SPARQL for sensor monitoring. Heritage question answering, however, targets static knowledge, like intangible heritage~\cite{xu2024nanjing}, museum content~\cite{vasic2025knowledge}, and numerical aggregation over genealogical graphs~\cite{suissa2023around}, not live sensor data.

\textbf{LLMs over structured data.} LLMs struggle with precise computation over tables~\cite{liu2024rethinking}. A common remedy has the LLM write code that an engine runs, such as table operations in Chain-of-Table~\cite{wang2024chain} or SQL in MAC-SQL~\cite{wang2025mac}, and TableRAG~\cite{chen2024tablerag} scales this to million-cell tables by retrieving relevant columns and cells; their results are only as correct as that code. PlanCompiler~\cite{harikumar2026plancompiler} instead compiles a validated, typed LLM plan deterministically, but for generic data-processing workflows.

\textbf{Knowledge graph question answering.} LLM-based KGQA~\cite{pan2024unifying} either reasons over the graph, as in Think-on-Graph~\cite{sun2024think}, or generates executable queries, as in Interactive-KBQA~\cite{xiong2024interactive}, schema-aware SAGA~\cite{zhang2026saga}, and a data-centre telemetry chatbot~\cite{khan2025data}. These methods return entities, timestamps, or the results of queries the LLM writes, rather than the spatial, temporal, and rule-based facts that HeriKGQA computes deterministically from observations.

%% file: sections/7_conclusion.tex
\section{Conclusion}

We presented HeriKGQA along with a heritage sensor knowledge graph and a QA benchmark with validated reference answers. Its core principle is that \textit{the LLM neither computes nor writes the code that computes}: it parses questions into constrained intents and verbalises grounded values, while compiled SPARQL and pre-implemented functions compute every number. HeriKGQA is more accurate than all baselines and ablations, with comparable answer quality and low latency, most clearly on spatial and anomaly-related questions, and the principle transfers to other domains through a new ontology mapping and rule implementations.

\section*{Limitations}
While HeriKGQA answers heritage sensor questions accurately, our work has five main limitations.

\textbf{Benchmark coverage.} Our 500-question benchmark covers five question types, but its questions are generated from templates over a single snapshot. Questions that real users asked during deployment are phrased more variably, combine several constraints, or rest on false premises, which the templates do not capture. Our evaluation on 80 user questions (Section~\ref{sec:deployment}) only partly addresses this, and we plan to extend the benchmark with user questions in future work.

\textbf{Graph-specific grounding layer.} Although the principle transfers to other datasets such as MultiTQ, the temporal, spatial, and anomaly computations are written for this knowledge graph: they encode which classes carry timestamps, which properties hold coordinates, and the anomaly rules, re-implemented from their original definitions. The parser and the SPARQL compiler read the schema from the ontology, so new classes and properties can be queried without code changes, but a new type of analysis or rule set requires new code.

\textbf{In-context examples.}
The parser's 14 examples, none from the benchmark, cover existing question types, whereas baselines receive at most three demonstrations (Table~\ref{tab:systems}). With prompts fixed, grounding accounts for 12.2 of the 18.2-point margin over Interactive-KBQA; the remaining 6.0 may partly reflect both the in-context examples and the intent parsing, neither of which we ablated.

\textbf{Large results are declined.} Questions whose listings exceed 1{,}000 rows are declined rather than truncated, to avoid answers computed from a sample. Questions that genuinely need many rows, such as listing all readings in a long window, are therefore outside the system's scope.

\textbf{Human evaluation.} The human evaluation compares HeriKGQA only with Interactive-KBQA on 100 sampled answer pairs, rated by three evaluators. Agreement is low (Fleiss' $\kappa = 0.129$), partly because the evaluators used ``tie'' at very different rates, and because answers from two systems using the same LLM are often hard to distinguish.

\section*{Ethics Statement}

\textbf{Use of AI assistants.}
Beyond the LLMs evaluated and used as system components in this work, we used generative AI assistants to polish the writing and help draft code. The research ideas and system design are the authors' own; the authors reviewed and verified all AI-assisted text and code and take full responsibility for the paper's content.

\textbf{Human evaluation.} Answer quality was compared by three volunteer evaluators with mixed backgrounds in cultural heritage and musicology. One of the three evaluators is a co-author. Evaluators were informed of the study's purpose and saw only questions and anonymised system outputs, and no personal information was collected about them beyond their academic background.

\textbf{Data and deployment.} The sensor data are open to the public under the project policy. In the deployment, questions, answers, and ratings are logged anonymously for research, as stated in the interface, without recording users' personal information.

\textbf{Risks.} Wrong answers could mislead conservation decisions. HeriKGQA mitigates this by computing every number deterministically, declining unanswerable questions, and exposing the time range, intent, and SPARQL query behind each answer, to support rather than replace expert judgement. LLM calls \revise{go to an institutional endpoint}, so no monitoring data reach third parties.

%% file: sections/appendix.tex
\section{Data and Benchmark}
\label{app:benchmark}
\textbf{Data.} Offline experiments use a frozen snapshot of the sensor database from 13~May to 12~August 2026 (Table~\ref{tab:data}(a)): 71 sensor nodes at five heritage sites host 113 sensors reporting 137 distinct observed properties, giving 1{,}677{,}556 readings and 5{,}227{,}617 observations. Each observation carries spatial context (the latitude, longitude, and altitude of its node, known for 68 of the 71 nodes) and temporal context (its timestamp), and is reported on a raw device channel such as \texttt{temperature} or \texttt{voltage\_1}. Of the 37 anomaly rules (Table~\ref{tab:data}(b)), 19 have usable readings at their site. Of the other 18, 12 refer to quantities that no sensor at that site reports, such as crack width, tilt, or indoor humidity; 5 refer to quantities reported, but not in the form the rule uses: rainfall as intensities and running totals, not windowed amounts (three rules, \siteA{} and \siteB{}), humidity on channels \siteA{}'s rule excludes, and GNSS as raw fixes, not long-term series; and one humidity rule (\siteB{}) needs smoothing not reproducible on irregular readings.

\textbf{Questions.} We wrote 52 templates covering \textit{basic} (counts and statistics of a quantity), \textit{spatial} (node positions, distances, and centroids), \textit{temporal} (time spans, daily or monthly statistics, and relative windows), \textit{anomaly-related} (rule verdicts and thresholds), and \textit{follow-up} questions (e.g., \texttt{``What about \siteB{}?''}), each preceded by one earlier turn (88 questions) or two (12). Spatial questions include \texttt{``Which two nodes at \siteA{} are closest together?''}, where distances matter at scale (the 19 nodes at \siteB{} span up to 1{,}670\,m), and temporal questions include \texttt{``What time period do the readings at \siteA{} cover?''}. Templates are instantiated with entities from the snapshot and sampled with a fixed seed and per-template quotas, giving 100 questions per type and 95--124 per site; questions coinciding with any system's in-context examples are excluded. No template was held out, so the benchmark measures accuracy on these question forms; generalisation to unseen formulations is evaluated separately on real user questions (Section~\ref{sec:deployment}) and on MultiTQ (Appendix~\ref{app:robustness}). Seven questions cannot be answered from $\mathcal{G}$, as they ask for unmeasured quantities (e.g., CO$_2$), unmodelled information, or periods outside the snapshot, and should be declined. Of the anomaly-related questions, 30 ask for a threshold and 70 for a verdict, balanced over \emph{Yes} (23), \emph{No} (25), and \emph{Cannot be determined} (22). Table~\ref{tab:examples} shows example questions with reference answers, including undeterminable and unanswerable ones.

\begin{table}[t]
\centering
\scriptsize{\textbf{(a) Data statistics}}\\[2pt]
\resizebox{\columnwidth}{!}{%
\begin{tabular}{lrrrrrrr}
\toprule
Site & Nodes & Sensors & \makecell[r]{Temperature\\sensors} & \makecell[r]{Observed\\properties} & Readings & Observations & Rules \\
\midrule
\siteA{}     & 15 & 30 & 19 & 67 &   709{,}170 & 2{,}138{,}510 &  8 \\
\siteB{}        & 19 & 30 & 10 & 43 &   225{,}945 &   580{,}873 &  4 \\
\siteC{}    & 15 & 20 & 18 & 39 &   574{,}578 & 1{,}454{,}889 & 16 \\
\siteD{}     & 13 & 19 &  9 & 54 &   108{,}150 &   511{,}970 &  5 \\
\siteE{} &  9 & 14 & 12 & 55 &    59{,}713 &   541{,}375 &  4 \\
\midrule
Total                  & 71 & 113 & 68 & 137$^\dagger$ & 1{,}677{,}556 & 5{,}227{,}617 & 37 \\
\bottomrule
\end{tabular}%
}

\vspace{8pt}
\scriptsize{\textbf{(b) Expert anomaly rules}}\\[2pt]
\resizebox{\columnwidth}{!}{%
\begin{tabular}{lllll}
\toprule
Site & Rule & Checks & Threshold(s) & Data \\
\midrule
\siteA{} & \texttt{R1\_ppv} & peak particle velocity (PPV) & 0.25 &  \\
 & \texttt{R2\_hum\_daily\_range} & daily humidity range & 10 &  \\
 & \texttt{R3\_gnss\_change} & day-over-day GNSS position change & 3 &  \\
 & \texttt{R4a\_soil\_high} & high soil moisture & 70 &  \\
 & \texttt{R4b\_soil\_range} & daily soil moisture range & 3 &  \\
 & \texttt{R5a\_airflow\_const} & airflow constancy & 0.05 & \checkmark \\
 & \texttt{R5b\_airflow\_range} & airflow range & 0.35 & \checkmark \\
 & \texttt{R6\_rain\_1h} & 1-hour rainfall & 10 &  \\
\midrule
\siteB{} & \texttt{R1a\_rain\_1h} & 1-hour rainfall & 10 &  \\
 & \texttt{R1b\_rain\_3h} & 3-hour rainfall & 20 &  \\
 & \texttt{R2\_wind} & wind speed & 12 & \checkmark \\
 & \texttt{R3\_hum\_cross} & humidity threshold crossings & 9; 75 & \\
\midrule
\siteC{} & \texttt{R10\_solar\_zscore} & daily GNSS z-score & 1 &  \\
 & \texttt{R1a\_crack\_step} & crack step & 0.7 &  \\
 & \texttt{R1b\_crack\_offset} & crack offset & 0.9 &  \\
 & \texttt{R2a\_tilt\_step} & tilt step & 0.7 &  \\
 & \texttt{R2b\_tilt\_offset} & tilt offset & 0.9 &  \\
 & \texttt{R3\_vib\_zscore} & vibration z-score & 7 &  \\
 & \texttt{R4a\_hum\_spike\_6h} & 6-hour humidity spike & 98 & \checkmark \\
 & \texttt{R4b\_hum\_persist\_7d} & 7-day humidity persistence & 90 & \checkmark \\
 & \texttt{R4c\_hum\_rapid\_1h} & rapid 1-hour humidity change & 20 & \checkmark \\
 & \texttt{R5\_visitors} & visitor peak & 3 & \checkmark \\
 & \texttt{R6a\_rain} & 1-hour and 24-hour rainfall & 5; 20 & \checkmark \\
 & \texttt{R6b\_rain\_cum7d} & 7-day cumulative rainfall & 100 & \checkmark \\
 & \texttt{R7\_wind} & wind peak ratio & 6 & \checkmark \\
 & \texttt{R8\_baro} & 6-hour barometric pressure range & 300 & \checkmark \\
 & \texttt{R9a\_temp\_lt} & daily temperature range & 20 & \checkmark \\
 & \texttt{R9b\_temp\_st} & short-term (1-hour) temperature change & 5 & \checkmark \\
\midrule
\siteD{} & \texttt{R1\_hum\_inside} & indoor humidity & 80 &  \\
 & \texttt{R2\_rain\_7d} & 7-day rainfall & 20 & \checkmark \\
 & \texttt{R3\_temp\_nodiff} & soil/air temperature difference & 2 &  \\
 & \texttt{R4a\_soil\_high} & high soil moisture & 60 & \checkmark \\
 & \texttt{R4b\_soil\_low} & low soil moisture & 5 & \checkmark \\
\midrule
\siteE{} & \texttt{R1\_temp\_hum} & low temperature with high humidity & 0; 90 & \checkmark \\
 & \texttt{R2\_crack\_step} & crack step & 0.1 &  \\
 & \texttt{R3\_vib\_zscore} & vibration z-score & 4.5 & \checkmark \\
 & \texttt{R4\_pga} & peak ground acceleration (PGA) & 0.5 & \checkmark \\
\bottomrule
\end{tabular}%
}
\caption{Statistics of the evaluation snapshot (13 May--12 Aug 2026), per site. (a) Data statistics: \emph{Readings} are \texttt{sosa:ObservationCollection}s and \emph{Observations} are \texttt{sosa:Observation}s; \emph{Rules} are \texttt{Policy} instances linking each rule to its site. $^\dagger$Distinct across sites, since sites share property names. (b) The 37 AD-RT anomaly rules with their thresholds; \checkmark: the snapshot contains usable readings the rule applies to.}
\vspace{-1em}
\label{tab:data}
\end{table}

\textbf{Reference answers.} References are computed from the snapshot in separate SQL and Python code that follows the rule conventions of Appendix~\ref{app:implementation}. Times are in UTC, relative windows end at the last observation (12~August 2026, 09:15), distances are haversine distances on a sphere of radius 6{,}371\,km (between nodes, also accepted with their altitude difference), and centroids are mean node coordinates. For anomaly questions, each (sensor, quantity) series is evaluated separately over windows that end at the rule's latest matching reading on or before the named day (or the snapshot's end), and series without readings in a window are skipped. A z-score rule compares each series' latest reading up to the end of that day with the mean and standard deviation of the series' readings up to then; as no recency is required, a series that has stopped reporting contributes its last reading; 4 \emph{Yes} verdicts at \siteE{} rest on stale readings (up to 29 days old). The verdict is \emph{Yes} if any applicable rule fires, \emph{No} if every applicable rule was evaluated and none fired, and \emph{Cannot be determined} otherwise. The rules of a question share their input quantity and are either all evaluated or none is, so no verdict rests on a subset of its rules. For the same reason, this definition coincides on the benchmark with the one in the baselines' prompts (Appendix~\ref{app:prompts}), under which \emph{No} requires at least one evaluated rule and none fired. A second, independent SQL implementation, which also recomputes earlier follow-up turns, reproduces the reference for all 489 questions whose answer can be computed, including the 2 unanswerable questions about unmeasured quantities, for which it finds no data; there is no disagreement. The other 11 (the 5 unanswerable questions about unmodelled information or periods outside the snapshot, and 6 anomaly questions with no applicable or evaluable rule) were adjudicated manually. Co-authors and colleagues from the pilot sites validated all references. Both implementations, per-question verification queries and decisions, and the matching rules (numerical tolerance, units, timestamps, lists) are in our repository.

\begin{table}[t]
\centering
\resizebox{\columnwidth}{!}{%
\begin{tabular}{p{2.2cm}p{5.4cm}p{3.4cm}}
\toprule
Type & Question & Reference answer \\
\midrule
Basic & What is the average soil temperature observed at \siteA{}? & 28.216 \\
Spatial & What is the distance between sensor nodes AF3 and GNSS-Rover at \siteA{}? & 11.0\,m \\
Temporal & What was the average temperature at \siteA{} over the last 24 hours? & 25.871 (4,713 observations, window ending 12~Aug, 09:15) \\
Anomaly & According to the AD-RT anomaly rules, was there a rainfall anomaly at \siteD{} on 2026-07-20? & No: 7-day rainfall 5.2, below the threshold of 20 (\texttt{R2\_rain\_7d}) \\
Follow-up & \textit{Q1:} What is the average temperature at \siteC{}? \textit{(23.639\,\textdegree C)} \newline
            \textit{Q2:} What about \siteA{}? \textit{(21.722\,\textdegree C)} \newline
            \textit{Q3:} And the maximum there? & 44.6\,\textdegree C \\
\midrule
Undeterminable (anomaly) & Did the crack width at \siteE{} exceed its AD-RT threshold on 2026-06-02? & Cannot be determined: no crack width readings at \siteE{} \\
Unanswerable & What is the average CO$_2$ concentration at \siteC{}? & Declined: no CO$_2$ data \\
\bottomrule
\end{tabular}%
}
\caption{Example benchmark questions. In the follow-up, earlier turns are shown with their reference answers.}
\label{tab:examples}
\end{table}

\section{Implementation Details}
\label{app:implementation}

\textbf{Knowledge graph mapping.} Each \texttt{sosa:ObservationCollection} is linked to the sensor that produced it, the node hosting that sensor, and the site where the node is deployed, so that every reading can be traced from its value to its site. Each anomaly rule is mapped to an instance of \texttt{Policy}, the PANOPTES class for conservation policies, which links the rule's definition and site-specific threshold to the site it applies to. Beyond policies, PANOPTES also models threats and conservation decisions.

\textbf{Parsing and compilation.}
The parser's system message contains the intent schema, an operator guide, the main graph paths, the rules that keep computation out of the parser, and 14 few-shot examples, followed by the schema of instantiated ontology classes, the labels of all observed properties, sites, and nodes, and each site's rule identifiers, so that filters use the graph's own names (e.g., \texttt{soil\_temperature}). The user message holds the reference time (the latest observation in $\mathcal{G}$), the last $h=2$ question--intent pairs with result summaries, the question, and, on a retry, the error, failed query, and previous intent. An intent is a JSON object with \texttt{target\_class}, \texttt{filters}, \texttt{return\_properties}, \texttt{joins}, \texttt{aggregation} (\texttt{count}, \texttt{min}, \texttt{max}, \texttt{avg}, \texttt{sum}, or \texttt{range}, optionally grouped by a property or by day or month), \texttt{order\_by}, \texttt{limit}, \texttt{anomaly\_check}, \texttt{as\_of}, \texttt{relative\_time}, and \texttt{unanswerable}, a reason for declining without a query. Invalid intents (e.g., contradictory filters) and non-JSON outputs trigger a retry. Compilation maps names to ontology identifiers: the target class becomes a type triple, return properties and unfiltered joins become \texttt{OPTIONAL} patterns, filters become \texttt{FILTER} expressions (a bare date denoting the whole UTC day), relative windows become timestamp filters ending at the reference time, and aggregations become aggregate functions with \texttt{GROUP BY}. Every aggregate also returns the number of aggregated entities (\texttt{n\_values}), and a result with \texttt{n\_values}$\,=0$ is treated as empty, so that an average over no data is not reported as 0. A limit combined with joins compiles into a subquery, so that ``the highest node at \siteA{}'' ranks only \siteA{}'s nodes. For example, ``What was the highest temperature at \siteA{} in the last 24 hours?'' compiles to:

\begin{small}
\begin{verbatim}
SELECT (MAX(?v) AS ?max)
       (COUNT(DISTINCT ?obs) AS ?n_values)
WHERE {
  ?oc a sosa:ObservationCollection ;
      sosa:resultTime ?t .
  FILTER(STR(?t) > "2026-08-11T09:15:00+00:00")
  ?oc herikg:hasObservation ?obs .
  ?obs sosa:observedProperty ?p .
  ?p rdfs:label ?pl .
  FILTER(LCASE(STR(?pl)) = LCASE("temperature"))
  ?oc herikg:hasSite ?site .
  ?site rdfs:label ?sl .
  FILTER(LCASE(STR(?sl)) = LCASE("Baltanas"))
  ?obs sosa:hasSimpleResult ?v . }
\end{verbatim}
\end{small}

\noindent Timestamps are rendered as UTC ISO 8601 strings with offset +00:00, so lexical comparison matches chronological order; no upper bound is needed, as the reference time is the latest observation.

\textbf{Follow-up resolution and execution details.} The parser runs at temperature~0, which also yields the greedy parse used in intent voting. For follow-ups, each result summary contains the row count, column names, and up to eight sample values; a newly named site or quantity replaces the previous one, and ``among these'' inherits the previous target and filters. Relative windows that lie entirely outside the data are declined, and the 1{,}000-row cap applies only to rows passed to the answer LLM, not to rule readings. Queries run on Ontop~5.5 over the read-only PostgreSQL snapshot with a 120\,s timeout; URIs are resolved to labels in batches of 50, as Ontop rejects batches of 100, and non-aggregated results above 1{,}000 rows are declined.

\textbf{Grounding and answer generation.}
Beyond the facts of Section~\ref{sec:method-layer}, temporal facts are measured from the latest observation in $\mathcal{G}$, and spatial facts add all pairwise distances for results of at most 25 nodes, per-site centroids for multi-site results, and facts about the nodes named in the question. For an anomaly question, the rules of the named site that match the named quantity are retrieved; running totals (cumulative rainfall and visitor counts) and rainfall intensity are never matched, as rule thresholds refer to amounts within a window. \emph{Direct} rules compare the latest reading with the threshold, \emph{windowed} rules compute a sum, mean, or range over 1\,h to 7\,d, \emph{z-score} rules compare the latest value with the mean and standard deviation of the sensor's values up to the cutoff, and \emph{custom} rules compute metrics such as the rate of change or a combination of two quantities; series are combined by their maximum, or minimum for rules that fire on low values or variability. The question's verdict follows the definition in Appendix~\ref{app:benchmark}. The answer LLM additionally receives the executed query and, for a follow-up, the earlier questions and the parser's interpretation; its prompt asks for one to three sentences that do not present a limited result as complete and avoid internal identifiers. Besides checking numbers, the verifier requires an anomaly answer to lead with the computed verdict; if the regenerated answer fails again, a template answer listing the rows and facts, or the verdict and evaluated rules, is returned. Every answer ends with the time range of the data it rests on.

\textbf{Rule implementations.}
The 19 evaluable rules are implemented once, for both the grounding layer and the rule definitions given to the baselines (Appendix~\ref{app:prompts}). A rule fires when its metric exceeds the threshold, or falls below it for \texttt{R4b\_soil\_low} and \texttt{R5a\_airflow\_const}, which detect low values or variability; \texttt{R1\_temp\_hum} fires when temperature falls below and humidity exceeds their thresholds. Windows end at the latest matching reading on or before the cutoff, and thresholds are stated in the channels' raw units. Eight rules approximate their expert definitions: \texttt{R5a\_airflow\_const}, \texttt{R5b\_airflow\_range}, and \texttt{R9a\_temp\_lt} use a rolling 24\,h window instead of calendar days; \texttt{R4c\_hum\_rapid\_1h} and \texttt{R9b\_temp\_st} take the largest change between consecutive readings in a rolling 1\,h window instead of AD-RT's resampled bins; \texttt{R4b\_hum\_persist\_7d} and \texttt{R6b\_rain\_cum7d} take a plain mean or sum over the raw readings of the 7-day window, the former instead of a mean of daily means; and \texttt{R1\_temp\_hum} pairs the latest reading of each quantity rather than readings at the same timestamp. Z-score rules use the population standard deviation of the series up to the cutoff. No rule requires a minimum number of readings or a recent one, so a series that has stopped reporting contributes its last value (Appendix~\ref{app:benchmark}). The \siteB{} humidity rule that requires smoothing is not implemented and is flagged without a verdict; the rule code is versioned in our repository.

\begin{table*}[t!]
\centering
\small
\renewcommand{\arraystretch}{1.2}
\setlength{\tabcolsep}{4pt}
\newcommand{\group}[1]{\multicolumn{5}{@{}l}{\textit{#1}}\\[1pt]}
\resizebox{\textwidth}{!}{%
\begin{tabular}{@{}>{\raggedright\arraybackslash}p{2.5cm}
  >{\raggedright\arraybackslash}p{3.8cm}
  >{\raggedright\arraybackslash}p{3.8cm}
  >{\raggedright\arraybackslash}p{3.8cm}
  >{\raggedright\arraybackslash\columncolor{gray!10}}p{3.8cm}@{}}
\toprule
 & \textbf{\textsc{TextRAG}} & \textbf{TableRAG} & \textbf{Interactive-KBQA} & \textbf{HeriKGQA} \\
\midrule
\group{Representation and querying}
Data representation & Plain-text rows, chunked by site, node, sensor, property, and day & One flattened table (pandas) & The same virtual RDF knowledge graph (Ontop) & Virtual RDF knowledge graph (Ontop) \\
Retrieval & Hybrid BM25 and dense retrieval (RRF) & LLM query expansion, then schema and cell retrieval (top 5 per query) & Multi-turn exploration via SPARQL tools & Single structured intent \\
Query generation & -- & Pandas commands written by the LLM in a ReAct loop & Free-form SPARQL written by the LLM & Deterministic compilation of the LLM's intent \\
Schema knowledge & Chunk titles only & Retrieved columns with types and example values & Hand-written schema, tool descriptions, and patterns & Ontology-derived schema of instantiated classes \\
In-context examples & None & No task-specific ones (official prompts) & 3 demonstration dialogs & 14 question--intent examples covering question types \\
\midrule
\group{Computation}
Statistics & LLM, over retrieved rows & LLM-written pandas code & LLM-written aggregates, if any & SPARQL aggregates, if parsed \\
Temporal and spatial facts & LLM inference & LLM-written pandas code or LLM inference & LLM inference or SPARQL & Deterministic grounding layer \\
Anomaly verdicts & LLM judgement against rule text & LLM judgement against rule rows, possibly via code & LLM-written threshold query & Per-sensor rule evaluation with site thresholds \\
Relative time & Data range stated in the prompt & Date range stated in the table caption & Inferred by the LLM, or queried as the latest timestamp & Anchored to the latest reading, timestamp reported \\
\midrule
\group{Robustness}
Large results & Truncation at 1,000 rows & Cell retrieval capped at 10,000 values; code runs on the full table & Truncation at 2,000 characters & Request to narrow above 1,000 rows, no truncation \\
Query failures & -- & LLM self-correction from execution errors (at most 5 steps) & LLM self-correction until an action repeats & Parser retry with error feedback (at most 2) \\
Follow-up questions & Previous turns in retrieval query and prompt & Earlier questions in the prompt & Earlier questions in the prompt & Short history, with new sites or quantities replacing old ones \\
\midrule
\group{Cost and observed behaviour}
LLM calls per question & 1 & 2 query expansions, plus up to 5 solver steps & Up to 10 steps $\times$ 3 candidates, plus 1 or 2 & 3 parses and 1 answer, plus up to 2 retries and 1 regeneration \\
Latency & Low, stable & Moderate & High, variable & Moderate \\
Main failure modes & Incomplete retrieval, arithmetic errors & Step limit reached, incorrect pandas code, false abstentions & Parsing errors, exploration detours, incorrect SPARQL & Intent parsing errors, unsupported question types \\
\bottomrule
\end{tabular}%
}
\caption{Comparison of the evaluated systems.}
\label{tab:systems}
\end{table*}

\begin{figure*}[!t]
\begin{lstlisting}[style=prompt]
[System Prompt]
You are a helpful and precise assistant in checking the quality of the answer.

[Instruction]
We would like to request your feedback on the performance of the response of the assistant to the user's questions displayed below. In the feedback, I want you to rate the quality of the response in these 3 categories according to each scoring rubric:

Readability: Is the response structured to promote readability and coherence? Does the response exhibit excellent organization? 
[Rubrics for scores 1-5 omitted]
Conciseness: Is the response presented in a concise manner for the reader without any unnecessary information? 
[Rubrics for scores 1-5 omitted]
Comprehension: Does the response fulfill the requirements of the questions by providing relevant information especially when the questions are complex and include multiple requirements? This includes responding in accordance with the explicit and implicit purpose of the given questions. 
[Rubrics for scores 1-5 omitted]

[User's Questions]
Earlier questions in this conversation:
- What was the average temperature at Baltanas on 2026-08-04?

Current question: And on the following day?

[Assistant's Responses]
The average temperature at Baltanas on 2026-08-05 was 23.10 degrees, based on 4,714 temperature observations.

[Instruction]
Please give feedback on the assistant's responses. Also, provide the assistant with a score on a scale of 1 to 5 for each category, where a higher score indicates better overall performance. Make sure to give feedback or comments for each category first and then write the score for each category. Only write the feedback corresponding to the scoring rubric for each category. The scores of each category should be orthogonal, indicating that 'correctness' should not be considered for the 'readability' category, for example.

Lastly, return a Python dictionary object that has category names as keys and the corresponding scores as values.
\end{lstlisting}
\caption{LLM-as-judge prompt (adapted from FLASK's prompt~\citep{ye2024flask} without sample answer; rubrics omitted). Only the questions and responses vary; for a question without earlier turns, only the current one is shown.}
\label{list:judge-prompt}
\end{figure*}

\begin{table*}[t]
\centering
\small
\setlength{\tabcolsep}{3pt}
\resizebox{\textwidth}{!}{%
\begin{tabular}{@{}l>{\columncolor{gray!12}}c@{\hspace{12pt}}ccccc@{\hspace{12pt}}ccc@{}}
\toprule
 & & \multicolumn{5}{c}{Question type} & \multicolumn{3}{c}{Category} \\
\cmidrule(lr){3-7} \cmidrule(l){8-10}
System & Mean & Basic & Spatial & Temporal & Anomaly & Follow & Readability & Conciseness & Comprehension \\
\midrule
TextRAG                    & 4.72 & 4.61 & 4.70 & 4.60 & 4.89 & 4.77 & 4.87 & 4.48 & 4.80 \\
TableRAG                   & 4.39 & 4.70 & 4.28 & 4.58 & 3.94 & 4.45 & 4.31 & \textbf{4.86} & 4.00 \\
Interactive-KBQA           & \textbf{4.85} & \textbf{4.93} & 4.84 & \textbf{4.80} & 4.95 & 4.71 & 4.98 & 4.72 & 4.84 \\
\midrule
HeriKGQA                   & 4.79 & 4.77 & 4.85 & 4.65 & \textbf{4.96} & 4.72 & 4.90 & 4.57 & \textbf{4.90} \\
\quad w/o grounding        & 4.78 & 4.82 & \textbf{4.90} & 4.72 & 4.70 & 4.76 & \textbf{5.00} & 4.63 & 4.71 \\
\quad w/o error correction & 4.82 & 4.82 & 4.86 & 4.67 & 4.94 & \textbf{4.81} & 4.99 & 4.57 & \textbf{4.90} \\
\bottomrule
\end{tabular}%
}
\caption{Answer quality by question type and category, rated by the LLM judge (1--5). Each type score averages the three category scores; \emph{Mean} (shaded) averages the five types. Best value per column in bold.}
\label{tab:quality-by-type}
\end{table*}

\begin{table}[t]
\centering
\setlength{\tabcolsep}{4pt}
\resizebox{\columnwidth}{!}{%
\begin{tabular}{@{}lrrrrr@{}}
\toprule
& & \multicolumn{3}{c}{Human preference} & \\
\cmidrule(lr){3-5}
& $n$ & HeriKGQA & IKBQA & Tie & \makecell{Agree w/\\judge} \\
\midrule
\multicolumn{6}{@{}l}{\textit{By judge verdict}} \\
\quad HeriKGQA win         &  34 & 31 &  1 &  2 & 91\% \\
\quad IKBQA win &  33 & 10 & 22 &  1 & 67\% \\
\quad Tie                     &  33 & 25 &  5 &  3 &  9\% \\
\midrule
\multicolumn{6}{@{}l}{\textit{By evaluator}} \\
\quad Evaluator 1             & 100 & 67 & 30 &  3 & -- \\
\quad Evaluator 2             & 100 & 51 & 31 & 18 & -- \\
\quad Evaluator 3             & 100 & 42 & 30 & 28 & -- \\
\bottomrule
\end{tabular}%
}
\caption{Human pairwise evaluation of HeriKGQA versus IKBQA (Interactive-KBQA) on 100 questions sampled evenly across judge verdicts (human label: majority vote). Inter-evaluator agreement is slight (Fleiss' $\kappa = 0.129$; observed 48.0\%, chance 40.3\%): all three agree on 23\% of questions and at least two on 98\%. Answers come from an earlier run (Appendix~\ref{app:quality}).}
\label{tab:human-eval}
\end{table}

\textbf{Settings and ablations.}
All results come from a single run per system. The parser votes over $n=3$ intents, the greedy parse and two sampled at temperature 0.7, and is retried at most $k=2$ times; the answer LLM runs at temperature 0.3. All systems use \texttt{gpt-oss-120b} through an OpenAI-compatible endpoint (three concurrent requests, exponential backoff, 120\,s timeout) and \texttt{qwen3-embedding-8b} for embeddings. Systems run one after another on the same endpoint to make latencies comparable. The \emph{w/o grounding} ablation computes no facts; for anomaly questions, each rule's windowed readings and definition are passed to the answer LLM, which must reach the verdict itself, and these readings count towards the 1,000-row limit. The answer check, defined against the computed facts, is skipped. The \emph{w/o error correction} ablation uses a single greedy parse, no retries, and no answer check, but an anomaly answer not leading with the computed verdict is still replaced by the template.

\textbf{Baselines.}
All baselines use the same LLM, receive the earlier questions of a follow-up, and are instructed to lead anomaly answers with the verdict and to say when the data cannot answer a question; Table~\ref{tab:systems} compares them. TextRAG serialises observations as text rows chunked by site, node, sensor, observed property, and day, with separate chunks for sites, nodes, rules, threats, and decisions; it retrieves chunks with BM25 and dense retrieval fused by reciprocal rank fusion, up to 1,000 rows or 50 chunks, and answers in one call. TableRAG~\citep{chen2024tablerag}, designed for million-cell tables, runs the official agent with its default settings (query expansion, schema and cell retrieval with $K=5$ and $B=10{,}000$, and a ReAct solver executing pandas commands for at most five steps) over the snapshot flattened into one table, with one row per observation, one per reading without observations, and one per rule; we raise the per-step output limit from 128 to 4,096 tokens, as the model reasons before answering. We verified on these runs that final answers are parsed as in the official code. Interactive-KBQA~\citep{xiong2024interactive} keeps its three tools (\textsc{SearchNodes}, \textsc{SearchGraphPatterns}, and \textsc{ExecuteSPARQL}), reimplemented over our graph, with a tool description covering the schema and three graph-specific patterns and three demonstration dialogs. Each of at most ten steps keeps the most frequent of $n=3$ candidate actions sampled at temperature 0.7, matching HeriKGQA's intent voting, and a final call verbalises the dialog, since the original returns only query results.

\section{Answer Quality Measurement}
\label{app:quality}

\textbf{LLM-as-judge.} We rate \emph{answer quality} with the LLM-as-judge protocol of FLASK~\citep{ye2024flask}, which scores a response on fine-grained quality categories, each with its own 1--5 rubric. We reuse FLASK\footnote{\url{https://github.com/kaistAI/FLASK}}'s prompt, category rubrics and score parser and adapt only the choice of quality categories, the inputs, the judge and its output limit, which we raise because the judge reasons before scoring. We use the same three categories for every answer so that scores are comparable across systems and question types: \emph{Readability}, \emph{Conciseness} and \emph{Comprehension} (whether the response addresses what the question asks); categories that require the correct answer are excluded because correctness is measured by accuracy. The judge prompt (Figure~\ref{list:judge-prompt}) varies only in the questions and the responses; HeriKGQA's provenance note is removed so that it does not reveal the system. For a follow-up question (e.g., ``And on the following day?''), earlier questions of its conversation are listed before the current question. The judge is \texttt{deepseek-v4.1-flash} (552B parameters), run on an institutional LLM endpoint with FLASK's temperature of 0.2, unlike \texttt{gpt-oss-120b}, which all systems use. We report the mean score of each category and question type per system in Table~\ref{tab:quality-by-type}.

\textbf{Human evaluation.} The human evaluation was carried out on the answers of an earlier run of the same two systems on a previous version of the data snapshot, which contained fewer sensor measurements. As a diagnostic check of the LLM judge, three volunteer evaluators compare HeriKGQA with Interactive-KBQA pairwise, as pairwise preferences are more reliable than absolute ratings for human evaluators~\citep{zheng2023judging}. Since the judge scores each answer separately, we derive its pairwise verdict from the two answers' quality scores averaged over the three categories; scores differing by less than 0.1 are a tie. We select 100 of the 500 questions across the three verdicts: the 34 and 33 questions with the largest score gaps in favour of HeriKGQA and of Interactive-KBQA, respectively, and 33 random ties. For each question, evaluators independently choose the better answer or a tie, based on the judge's category descriptions (Figure~\ref{list:judge-prompt}). The answer order is randomised, and evaluators do not see the system names, the provenance note or the judge's verdict. The human label is the majority vote (a tie if all three disagree). Table~\ref{tab:human-eval} reports the results. Inter-evaluator agreement is slight (Fleiss' $\kappa = 0.129$), mainly because evaluators used ``tie'' at very different rates (3, 18, and 28 times), while all three prefer HeriKGQA more often than Interactive-KBQA. The human label matches the judge on 79\% of its non-tie verdicts (91\% when the judge favours HeriKGQA, 67\% when it favours Interactive-KBQA), likely an upper bound since these are the judge's clearest verdicts, but on only 9\% of its ties, where humans prefer HeriKGQA in 25 of 33 cases. Across these 100 questions, humans prefer HeriKGQA on 66 and Interactive-KBQA on 28.

\section{Failure Analysis and Case Studies}
\label{app:analysis}

\textbf{Failure taxonomy.} We assign each wrong answer to exactly one category, checked in the following order: (i) \emph{no answer} (an empty answer, or a reasoning or tool-call trace instead of an answer); (ii) \emph{declined (size)} (declined because the results exceed the 1{,}000-row cap); (iii) \emph{false abstention} (an answerable question answered with ``cannot be determined'' or ``no data''); (iv) \emph{missed decline} (an unanswerable question answered anyway); (v) \emph{wrong verdict} (a wrong anomaly verdict); and (vi) \emph{wrong value} (a wrong number, timestamp, node, or site, or a failed query). Table~\ref{tab:failures} gives the counts.

\begin{table}[t]
\centering
\setlength{\tabcolsep}{3.5pt}
\resizebox{\columnwidth}{!}{%
\begin{tabular}{@{}lrrrrrrr@{}}
\toprule
System & Wrong & \makecell{No\\answer} & \makecell{Declined\\(size)} & \makecell{False\\abst.} & \makecell{Missed\\decline} & \makecell{Wrong\\verdict} & \makecell{Wrong\\value} \\
\midrule
TextRAG               & 312 &  8 &  0 & 61 & 0 & 44 & 199 \\
TableRAG              & 259 & 92 &  0 & 53 & 1 & 25 &  88 \\
Interactive-KBQA      &  96 &  2 &  0 & 19 & 3 & 25 &  47 \\
\midrule
HeriKGQA              &   5 &  0 &  0 &  2 & 1 &  0 &   2 \\
\quad w/o grounding   &  66 &  0 & 35 &  2 & 1 &  5 &  23 \\
\quad w/o error corr. &  12 &  0 &  0 &  9 & 1 &  0 &   2 \\
\bottomrule
\end{tabular}%
}
\caption{Wrong answers out of 500 per system, by first matching failure category (in column order).}
\label{tab:failures}
\end{table}

\textbf{HeriKGQA's errors.} None of HeriKGQA's five errors comes from arithmetic. Four start in the parser and one in query execution. Two are false abstentions. In A071, the parser reads ``1-hour and 24-hour rainfall rule'' as a rule the graph lacks, because the rule's name (``rainfall short window'') does not mention its two windows. In F099, a follow-up asking ``And the relative humidity?'' at \siteC{}, the parser declares relative humidity unmodelled, although in the five other relative-humidity questions it maps the quantity to the weather station's humidity channel. In the missed decline (B099), the parser maps ``solar radiation'' to the illuminance channel (measured in lux) and reports its mean instead of declining. In T099, the parser omits the aggregation for ``What was the relative humidity at \siteA{} on 2026-07-01?''. The query therefore returns the 45 distinct values rather than all 296 readings, and their mean ($55.2\%$) differs from the reference ($57.1\%$). In T025, the compiled ranking query for the highest peak ground acceleration at \siteE{} exceeds the 120\,s timeout, and each re-parse yields an equivalent query that times out as well. Rule retrieval matches the quantity by keyword and can also retrieve an unrelated rule: in A061, ``wind'' also matches the description of a rainfall rule (``short-window rainfall''), which is evaluated and found within its threshold. As \emph{No} requires every retrieved rule to be evaluated, and the actual wind rule has no readings up to that day, HeriKGQA still answers \emph{Cannot be determined}, as the reference does; an unrelated rule that fired, however, would still yield a wrong \emph{Yes}.

\textbf{Error correction.} Without error correction, 7 of the 12 errors follow an empty query result. In B094 and F025, a misplaced join (a reversed node join; a value attached to the reading rather than its observation) returns no rows. In B098, F098, F100 and F036, a property label is not in the graph (``humidity'' instead of the channel \texttt{hum1}, and ``soil water potential'' instead of \texttt{soil\_water\_potential}). With error correction, 6 of these 7 questions are answered correctly, either because the empty result is fed back to the parser for a retry or because voting selects a parse whose query returns rows. In the seventh, T099, voting selects such a parse, so no retry is triggered, but that parse omits the aggregation (see above). Overall, 12 questions need at least one retry and 11 of them are then answered correctly; the exception is T025, where every query times out. The answer check triggers a regeneration for 107 of the 500 answers, and 55 of these are replaced by the deterministic template answer.

\textbf{Baselines.} Interactive-KBQA mostly computes wrong values (47) or verdicts (25), and 12 of its 19 false abstentions come without a single query that returned rows. TextRAG's errors are dominated by wrong values (199 of 312), since aggregates cannot be computed from a sample of rows, and 61 of its answers wrongly declare the retrieved data insufficient. Of TableRAG's 500 runs, 352 end with a final answer and 148 exhaust the five ReAct steps, in which case the final step is taken as the answer; 92 of these 148 yield no answer, as the final step is reasoning or a pandas command.

\textbf{Why anomaly accuracy drops without grounding.}
Without grounding, HeriKGQA answers 66 of the 100 anomaly questions correctly instead of 99, and its 34 errors have two causes. In 29 questions, all of them verdicts at \siteC{} and \siteE{}, the raw rule readings passed to the LLM exceed the 1{,}000-row cap (up to 75{,}344 rows), so the system declines; with grounding, these readings are evaluated in code and never enter the prompt. Of the 41 verdict questions whose readings fit, the LLM decides 36 correctly, against 41 with grounding. Four of its errors answer \emph{No} where no matching readings exist, treating the absence of an exceedance as evidence that none occurred, and one pools readings across sensors (Case~1). Threshold lookups, which need no computation, are unaffected (30/30).

\textbf{Case 1: pooled sensors cause a false alarm (A005).}
For ``According to the AD-RT expert rules, was there a temperature anomaly at \siteC{} on 2026-05-14?'', rule R9a fires when a sensor's daily temperature range exceeds 20\,$^\circ$C. The window holds 911 readings from 17 sensors, with a largest per-sensor range of 19.0\,$^\circ$C, so the reference is \emph{No}, as HeriKGQA answers. Without grounding, the LLM receives the same readings and the instruction to evaluate each sensor separately, yet answers \emph{Yes} with a range of 20.7\,$^\circ$C: the maximum of one sensor (22.7\,$^\circ$C) minus the minimum of another (2.0\,$^\circ$C). Interactive-KBQA makes the same mistake with the same numbers.

\textbf{Case 2: geodesic arithmetic (S005).}
For ``How far apart are the two most distant sensor nodes at \siteE{}?'', the reference is 5.3\,m, or 8.4\,m with the altitude difference; HeriKGQA obtains the former from pairwise haversine distances. Other systems estimate or misread it. Without grounding, the LLM picks the pair farthest apart with altitude but reports only their altitude difference (8.2\,m). Interactive-KBQA derives about 5.9\,m from the ranges of latitude and longitude, which no node pair spans. TextRAG picks the same pair but overstates its horizontal offset (5\,m instead of 1.6\,m, about 10\,m in total), and TableRAG answers ``0 km''. When nodes lie a few metres apart, approximate coordinate arithmetic is too imprecise.

\textbf{Case 3: a z-score rule needing the full series (F073).}
After ``Was the vibration at \siteE{} anomalous on 2026-07-23 according to its AD-RT rules?'', the user asks ``What about on 2026-07-27?''. Rule R3 compares each sensor's latest reading up to that day with the mean and standard deviation of its series. All systems resolve the follow-up, and HeriKGQA correctly answers \emph{Yes} (z-score 7.42). The baselines all answer \emph{Cannot be determined}: Interactive-KBQA queries only the wrong ground-acceleration series, ending in May; TextRAG retrieves no vibration readings; TableRAG filters the day's rows but computes nothing. None assembles the per-sensor series a z-score needs. Without grounding, HeriKGQA declines: the raw readings of the two vibration rules (63{,}041 rows across all sensors) exceed the 1{,}000-row cap.

\section{Additional Experiments}
\label{app:robustness}

\begin{table}[t]
\centering
\setlength{\tabcolsep}{4pt}
\resizebox{\columnwidth}{!}{%
\begin{tabular}{@{}l>{\columncolor{gray!12}}c@{\hspace{8pt}}cccccc@{}}
\toprule
\multicolumn{8}{@{}l}{\textit{(a) Generality: MultiTQ}} \\
\midrule
System & Mean & Equal & \makecell{Before/\\After} & \makecell{First/\\Last} & \makecell{Equal\\Multi} & \makecell{Before\\Last} & \makecell{After\\First} \\
\midrule
HeriKGQA         & \textbf{75.4} & \textbf{94.0} & 71.4 & \textbf{96.3} & 53.7 & \textbf{67.9} & \textbf{69.0} \\
Interactive-KBQA & 73.4 & 90.5 & \textbf{72.6} & 92.7 & \textbf{56.1} & 61.9 & 66.7 \\
\midrule
\addlinespace[2pt]
\multicolumn{8}{@{}l}{\textit{(b) Robustness: backbone LLM on our benchmark}} \\
\midrule
Backbone & Mean & Basic & Spatial & Temporal & Anomaly & Follow-up & \\
\midrule
\texttt{gpt-oss-120b} & \textbf{99.0} & \textbf{99} & \textbf{100} & 98 & 99 & \textbf{99} & \\
\texttt{qwen3-8-27b}  & \textbf{99.0} & 98 & 99 & \textbf{99} & \textbf{100} & \textbf{99} & \\
\bottomrule
\end{tabular}%
}
\caption{Robustness of HeriKGQA. (a) Accuracy (\%) on 500 MultiTQ questions sampled across its six question types (82--84 each), with both systems using \texttt{gpt-oss-120b}. (b) Accuracy (\%) of HeriKGQA with two LLM backbones on our 500 benchmark questions (100 per type) as in Table~\ref{tab:main-results}. \emph{Mean} averages question types. Best per panel in bold.}
\label{tab:multitq}
\label{tab:backbone}
\end{table}

\textbf{Generality on MultiTQ.} We also evaluate HeriKGQA and Interactive-KBQA on 500 test questions from the temporal KGQA benchmark MultiTQ~\cite{chen2023multi}, sampled across its six question types. Porting HeriKGQA replaces only its domain-facing parts (the ontology, prompts and 8 few-shot examples, an entity linker for the graph's ${\sim}$10.5k entities, and a script loading the data, about 500 lines in all), while the compiler and error correction are reused and the temporal grounding layer gains one generic extension, a time filter anchored at another event's time (under 100 lines); both systems use \texttt{gpt-oss-120b} and the same entity linker. HeriKGQA is competitive, with a modest advantage in this single run (75.4\% vs.\ 73.4\%; Table~\ref{tab:multitq}(a)): it gains on questions answered by an ordering over timestamps (e.g., Before Last, $+$6.0), which the compiler expresses in SPARQL, but trails slightly where a reference time must be resolved from another event, since it declines when that lookup fails. The design thus carries over to a different graph with competitive accuracy, and its advantage is clearest on heritage data, where spatial and rule-based computation matters most.

\textbf{Smaller backbone.} To test whether HeriKGQA depends on a large LLM, we rerun it on the 500 benchmark questions with \texttt{qwen3-8-27b}, a 27B-parameter model, in place of \texttt{gpt-oss-120b}, leaving the configuration, prompts, embeddings and snapshot unchanged. Accuracy does not drop ($99.0\%$ with both backbones; Table~\ref{tab:backbone}(b)): the two backbones differ on only 6 questions, 3 in each direction. Of the 3 questions that only Qwen gets wrong, two report the number of observations instead of readings and one names the wrong nearest node. Because the LLM only has to map a question to an intent, and every number in the answer is computed by SPARQL and the grounding layer, a smaller backbone is enough.

\lstdefinestyle{promptu}{basicstyle=\ttfamily\scriptsize, breaklines=true, breakatwhitespace=true,
    columns=flexible, frame=single, keepspaces=true, extendedchars=true, upquote=true,
    literate={→}{{$\rightarrow$}}1 {←}{{$\leftarrow$}}1 {↔}{{$\leftrightarrow$}}1 {—}{{\textemdash}}1
             {…}{{\ldots}}1 {Δ}{{$\Delta$}}1}
             
\section{LLM Prompts}
\label{app:prompts}

This appendix shows the main parts of the prompts sent to the LLM, rendered from the experiment code; omitted parts are marked with square brackets, and everything else is verbatim. None of the parser's 14 few-shot examples is a benchmark question; questions coinciding with any system's in-context examples were excluded (Appendix~\ref{app:benchmark}).

\textbf{Semantic parser (system message).}
The parser outputs an intent and computes nothing: statistics are computed in the query, and distances, times and rule verdicts by the grounding layer.
\begin{lstlisting}[style=promptu]
You are a semantic parser for a heritage sensor knowledge graph.

Given a natural-language query, output ONLY a valid JSON object (no markdown fences)
with this exact schema:

{
  "target_class": "<class name, one of: Asset, AssetState, Decision, Device, DeviceType, EngineeringUnit, InstalledDevice, InstalledInstrument, Instrument, InstrumentType, Location, Node, Observation, ObservationCollection, ObservedProperty, PhysicalQuantity, Policy, Sensor, Site, Threat>",
  "filters": [
    {"property": "<property>", "op": "contains|equals|in|gt|lt|from|until|after|before", "value": "<string, or a list for in>"}
  ],
  "return_properties": ["<list of properties to SELECT>"],
  "joins": [
    {
      "class": "<related class>",
      "via": "<property name>",
      "direction": "outgoing|incoming",
      "return_properties": ["<properties of the related class>"],
      "filters": [
        {"property": "<property>", "op": "contains|equals|in|gt|lt|from|until|after|before", "value": "<string, or a list for in>"}
      ]
    }
  ],
  "aggregation": null or {"type": "count|min|max|avg|sum|range", "group_by": "ClassName.property", "time_bucket": null|"day"|"month", "target": "<class or property>", "distinct": true|false},
  "order_by": null or {"variable": "ClassName.property", "direction": "ASC|DESC"},
  "limit": null or <integer>,
  "anomaly_check": true|false,
  "as_of": null or "<ISO date or datetime, e.g. '2025-02-11'>",
  "relative_time": null or {"amount": <integer>, "unit": "hours|days|weeks"},
  "unanswerable": null or "<what the knowledge graph does not model>",
  "explanation": "<one sentence>"
}

[Operator guide: contains, equals, gt/lt, in, from, until, after, before.]

Key graph paths (direct edges — use these for joins):
  ObservationCollection —[hasSite]→       Site
  ObservationCollection —[madeBySensor]→  Sensor
  ObservationCollection —[hasObservation]→ Observation
  ObservationCollection —[hasNode]→       Node
  Observation           —[observedProperty]→ ObservedProperty
[Multi-hop paths, variable naming, node coordinates, the governance graph of rules and pilot metadata.]

Rules:
[...]
  - "unanswerable": set (to a short reason) ONLY when the question needs something this knowledge graph does not model at all — e.g. ownership, costs, opening hours, construction materials, people — i.e. nothing in the schema or the label lists could hold it. Never for a quantity or period that might merely have no readings (that is decided by running the query), and never for an expert-rule/anomaly question (whether a rule applies is decided by the AD-RT rule evaluation). Leave the rest of the intent minimal when set.

Queries and aggregation (rules of the query compiler):
[...]
  - AGGREGATE FIRST — this is a correctness rule, not an optimisation. If the question asks for a statistic ("average", "mean", "total", "sum", "how many", "highest", "lowest", "maximum", "minimum"), you MUST set "aggregation" so the statistic is computed inside the query. Detail rows go through DISTINCT de-duplication, so any average, sum or count derived from them afterwards is WRONG. Only omit "aggregation" when the user genuinely wants the individual values listed, or when the statistic depends on distances between nodes (see the spatial rules: those are computed downstream from the coordinates).
[...]

Time (rules of the temporal layer):
  - Time periods are resultTime filters, written with bare dates and from/until, both ends inclusive:
[...]
  - "relative_time": set for a window that ends now ("in the last 24 hours", "over the past 7 days", "this past week" = 1 week). "Now" is the latest reading in the knowledge graph (given with the query as the reference time), not today's date; the window is turned into a resultTime filter downstream, so do NOT also add resultTime filters for it. Null otherwise.
[...]

Spatial questions (rules of the spatial grounding layer):
  - Counting or picking nodes BY DISTANCE ("how many nodes are within R m of X", "which node is closest to X") is a spatial question, not an aggregation: set aggregation to null and return the coordinates of all nodes of that site — the count or the nearest node is computed downstream from the distances. An aggregation here would count the site's nodes without any distance.
[...]

Expert rules and anomalies (rules of the AD-RT rule evaluation):
[...]
  - "anomaly_check": true only for a yes/no verdict against a known threshold ("has X exceeded threshold", "is there an anomaly"). False for a plain "what is the threshold for X" lookup, which wants the definition, not a verdict.
  - When anomaly_check is true: build the intent exactly like a Policy threshold lookup (target_class=Policy, filter by ruleId/label contains the quantity, join Site via appliesToSite) — do NOT join or aggregate Observation data yourself. A separate deterministic step (outside this JSON) takes the rule(s) this query returns and checks each against its threshold; it needs the rule identity first, which is what this query provides. Do not switch off target_class=Policy just because the question mentions a quantity/site.
[...]

Follow-up handling (earlier turns may appear above as question → intent pairs):
  - Treat the current question as SELF-CONTAINED by default. If it can be understood on its own, IGNORE the earlier turns completely.
  - When resolving, copy only the specific missing piece. NEVER carry over filters, aggregations, limits, joins or target_class that the current question does not itself imply. A new site/quantity mentioned now REPLACES the old one — it does not add to it.
  - Relative periods in a follow-up ("the following day", "the next month", "the day before") are relative to the period of the previous turn — read its date from the previous intent and shift it: previous "on 2025-03-04" + "the following day" → from 2025-03-05, until 2025-03-05.
[...]

--- EXAMPLES ---

Q: "Did the airflow at Baltanas trigger an expert-rule alarm on 2025-02-11?"
{
  "target_class": "Policy",
  "filters": [{"property": "ruleId", "op": "contains", "value": "airflow"}],
  "return_properties": ["policyDescription", "ruleId", "thresholdValue", "thresholdJson"],
  "joins": [
    {"class": "Site", "via": "appliesToSite", "direction": "outgoing",
     "return_properties": ["label"], "filters": [{"property": "label", "op": "equals", "value": "Baltanas"}]}
  ],
  "aggregation": null,
  "order_by": null, "limit": null,
  "anomaly_check": true,
  "as_of": "2025-02-11",
  "explanation": "Same Policy lookup as any anomaly_check question — the date is NOT a filter here (Policy has no resultTime), it goes in as_of so the deterministic step checks the rule against readings as of that day instead of the latest available."
}

[13 further examples: averages, values and sensors per site, the latest readings, a site's quantities, a monthly count, the month with the lowest mean, a 48-hour count, node coordinates, a threshold lookup, an anomaly check on the latest data, a follow-up, and pilot metadata.]

--- END EXAMPLES ---

[The ontology schema: 26 classes and their properties, generated from the ontology file.]

Labels in this knowledge graph — filter on them verbatim with op equals (or in), choosing the label that matches the user's wording (e.g. "soil temperature" → soil_temperature, "voltage" → voltage, not voltage_1). A quantity named in the question is ONE label: use equals, never contains (contains "voltage" also matches voltage_1 and voltage_polynomial_4). Use contains only when the user explicitly asks for a family ("all voltage channels"). A name that appears under Node.label is a sensor node, not a quantity.
[The 71 node labels, the 137 quantity labels, the 5 site labels, and the rule identifiers per site.]
\end{lstlisting}

\textbf{Semantic parser (user message).}
Earlier turns of a follow-up precede it as question--intent pairs, each with a note on the previous result (row count, columns and sample values).
\begin{lstlisting}[style=promptu]
[Reference time — the latest reading in the knowledge graph: 2026-08-12 09:15:00+00:00 UTC. Relative time expressions refer to this time, not to today's date.]
Query: <question>

On a retry, appended:
--- FEEDBACK FROM PREVIOUS ATTEMPT ---
<error, failed query and previous intent>
Please fix the intent JSON based on this feedback. Use only SHORT property names without namespace prefix. Output ONLY the corrected JSON.
\end{lstlisting}

\textbf{Answer generator (system message).}
For the w/o grounding ablation, the second sentence reads ``You are given a user question and SPARQL results. No numeric summary or analysis is provided.'', the LLM computes times, distances and ranges and receives rule definitions and readings, not verdicts.

\begin{lstlisting}[style=promptu]
You are a concise sensor data analyst for a cultural heritage monitoring system.

You are given a user question, SPARQL results, and a numeric summary.

Rules:
- Answer the question directly in 1-3 short sentences.
[...]
- If the question asks about something the results do not contain (something the knowledge graph does not model, e.g. ownership or costs), say it cannot be answered from the knowledge graph — do not answer a different question with the rows you have.
- Do NOT output URIs or internal identifiers — use human-readable names and labels only.
[...]

Numbers and facts:
- If the analysis has a section "Facts for this question", it was computed for exactly what was asked (a named pair's distance, the node nearest to a named node, the nodes within a radius, the distance between site centroids) — answer from it.
- Report numbers as the results/analysis give them, in the same unit (do not convert m to km or back) and with at least the decimals shown in the analysis (distances to 0.1 m or 0.1 km).

Statistics:
- When the query already computed an aggregate (a count/avg/min/max/sum column), report that value — it is the authoritative one.
- Otherwise the rows are a de-duplicated listing, so do NOT compute an average, total or record count from them; give the values, ranges or examples the rows support and leave it there.
[...]

AD-RT rule check (when present in the prompt):
- If a message includes a section starting "AD-RT rule check", that verdict (EXCEEDS/within a threshold) was computed in Python, not by you — report it as-is, do NOT recompute or second-guess it by comparing raw numbers yourself.
[...]
- If the prompt says "This is a yes/no anomaly check", START with "Yes."/"No."/"Cannot be determined." (Yes if ANY listed rule fired), then give the supporting numbers and, if multiple rules were checked, which fired.
\end{lstlisting}

\textbf{Answer generator (user message).}

\begin{lstlisting}[style=promptu]
[For a follow-up: the earlier questions and how the parser resolved the follow-up.]

User question:
<question>

SPARQL query executed:
```sparql
<compiled query>
```

Results — all <n> distinct rows (<m> total):
<rows as JSON>

Pre-computed spatial & temporal analysis:
<facts of the grounding layer>

<AD-RT rule check, for an anomaly question>

(This is a yes/no anomaly check — lead your answer with Yes/No/Cannot be determined.)

[After a failed answer check:]
--- YOUR PREVIOUS ANSWER FAILED A CHECK ---
<what was wrong>
Write a corrected answer.
\end{lstlisting}

An example of the rule check computed by the grounding layer (question A005):
\begin{lstlisting}[style=promptu]
AD-RT rule check for site='Lucretili', quantity~'R9a_temp_lt, R9b_temp_st', as of 2026-05-14T23:59:59 (computed here, not by the model):
  - R9a_temp_lt: daily range (max-min per day). Window(s) ending 2026-05-14T23:55:00+00:00: 24h range=19 (thr 20, max of 17 sensor series) → combined metric 19 vs threshold 20 → within. (Despite the '_lt' in the name this is a RANGE rule, not a less-than rule; approximated here as a rolling 24h window.)
  - R9b_temp_st: max absolute 1h change (rate of change). max |consecutive Δ| over 1h = 0.1, max of 6 sensor series vs threshold 5 → within. (approximated as the max |consecutive-sample difference| within a rolling real-time 1h window, not AD-RT's resample-bin-based rolling window)
\end{lstlisting}

\textbf{Instructions added to the baselines.}
All three baselines receive the earlier questions of a follow-up, the instruction to lead an anomaly answer with the verdict, and the instruction to say when the data cannot answer the question; TextRAG and Interactive-KBQA also receive the rule definitions, which TableRAG finds as rows of its table. Each rule definition follows the matcher and computation of HeriKGQA's rule evaluation, e.g.:
\begin{lstlisting}[style=promptu]
- site Lucretili: Rule R9a_temp_lt ('AD-RT Lucretili policy - daily temperature range'), threshold=20, threshold parameters={"R9a_temp_lt": 20}, quantities: label contains any of temp; label contains none of rain_acc, counter_acc, rain_1, definition: daily range (max-min per day), windows: 24h range vs threshold 'R9a_temp_lt', note: Despite the '_lt' in the name this is a RANGE rule, not a less-than rule; approximated here as a rolling 24h window.
\end{lstlisting}

TextRAG (system message; the rule definitions are added to the retrieved rule chunks):
\begin{lstlisting}[style=promptu]
You are a sensor data analyst for a cultural heritage monitoring system.
You are given a user question and sensor data retrieved for it, as plain-text rows.
Answer the question directly in 1-3 sentences, using only the retrieved data, and include the key numbers.
If the retrieved data does not contain the information needed, say that it cannot be determined from the data.
AD-RT expert-rule evaluation (how a verdict is decided): each rule is evaluated per data series — one quantity of one sensor — never on readings pooled across sensors or quantities; time windows end at the latest matching reading (at or before the end of the day the question names, if any); a rule fires if any series crosses its threshold. A question's verdict is Yes if any applicable rule fires, No if at least one rule could be evaluated and none fired, and Cannot be determined if no applicable rule exists or none could be evaluated (e.g. no matching readings).
If the question asks whether an anomaly occurred or a threshold was exceeded, start the answer with "Yes.", "No." or "Cannot be determined."
\end{lstlisting}

Interactive-KBQA (in addition to its original tool description and demonstrations):
\begin{lstlisting}[style=promptu]
[Appended to the tool description: the expert-rule evaluation paragraph given above for TextRAG, and one definition per rule, 37 in all.]

[Final request, after the dialog:]
Based only on the observations above, answer the question: <question, with earlier questions for a follow-up>
Answer directly in 1-3 sentences with the key numbers. If the observations do not contain the information needed, say that it cannot be determined from the knowledge graph.
If the question asks whether an anomaly occurred or a threshold was exceeded, start the answer with "Yes.", "No." or "Cannot be determined."

[Only if that answer was another tool call:]
The tools are no longer available, so no further query can be run. Do not write an Action. Give the final answer now, as asked above, based only on the observations already in this conversation.
\end{lstlisting}

TableRAG (the question passed to the agent):
\begin{lstlisting}[style=promptu]
<question> (If the question asks whether an anomaly occurred or a threshold was exceeded, start the final answer with Yes, No or Cannot be determined. If the table does not contain the information needed, the final answer should say that it cannot be determined.)
\end{lstlisting}

A follow-up question is passed to TextRAG, TableRAG, and Interactive-KBQA as:
\begin{lstlisting}[style=promptu]
Earlier questions in this conversation: "<earlier question>"; ... . Current question: <question>
\end{lstlisting}

%% file: custom.bib
@inproceedings{liu2024rethinking,
  author       = {Tianyang Liu and
                  Fei Wang and
                  Muhao Chen},
  title        = {Rethinking Tabular Data Understanding with Large Language Models},
  booktitle    = {{NAACL-HLT}},
  pages        = {450--482},
  publisher    = {Association for Computational Linguistics},
  year         = {2024},
  url          = {https://doi.org/10.18653/v1/2024.naacl-long.26},
  doi          = {10.18653/V1/2024.NAACL-LONG.26}
}

@inproceedings{wang2025mac,
  author       = {Bing Wang and
                  Changyu Ren and
                  Jian Yang and
                  Xinnian Liang and
                  Jiaqi Bai and
                  Linzheng Chai and
                  Zhao Yan and
                  Qian{-}Wen Zhang and
                  Di Yin and
                  Xing Sun and
                  Zhoujun Li},
  title        = {{MAC-SQL:} {A} Multi-Agent Collaborative Framework for Text-to-SQL},
  booktitle    = {{COLING}},
  pages        = {540--557},
  publisher    = {Association for Computational Linguistics},
  year         = {2025},
  url          = {https://aclanthology.org/2025.coling-main.36/}
}

@inproceedings{xiong2024interactive,
  author       = {Guanming Xiong and
                  Junwei Bao and
                  Wen Zhao},
  title        = {Interactive-KBQA: Multi-Turn Interactions for Knowledge Base Question
                  Answering with Large Language Models},
  booktitle    = {{ACL} {(1)}},
  pages        = {10561--10582},
  publisher    = {Association for Computational Linguistics},
  year         = {2024},
  url          = {https://doi.org/10.18653/v1/2024.acl-long.569},
  doi          = {10.18653/V1/2024.ACL-LONG.569}
}

@inproceedings{chen2024tablerag,
  author       = {Si{-}An Chen and
                  Lesly Miculicich and
                  Julian Eisenschlos and
                  Zifeng Wang and
                  Zilong Wang and
                  Yanfei Chen and
                  Yasuhisa Fujii and
                  Hsuan{-}Tien Lin and
                  Chen{-}Yu Lee and
                  Tomas Pfister},
  title        = {TableRAG: Million-Token Table Understanding with Language Models},
  booktitle    = {NeurIPS},
  year         = {2024},
  url          = {http://papers.nips.cc/paper\_files/paper/2024/hash/88dd7aa6979e352fda7c4952ca8eac59-Abstract-Conference.html}
}

@article{pan2024unifying,
  author       = {Shirui Pan and
                  Linhao Luo and
                  Yufei Wang and
                  Chen Chen and
                  Jiapu Wang and
                  Xindong Wu},
  title        = {Unifying Large Language Models and Knowledge Graphs: {A} Roadmap},
  journal      = {{IEEE} Trans. Knowl. Data Eng.},
  volume       = {36},
  number       = {7},
  pages        = {3580--3599},
  year         = {2024},
  url          = {https://doi.org/10.1109/TKDE.2024.3352100},
  doi          = {10.1109/TKDE.2024.3352100}
}

@inproceedings{sun2024think,
  author       = {Jiashuo Sun and
                  Chengjin Xu and
                  Lumingyuan Tang and
                  Saizhuo Wang and
                  Chen Lin and
                  Yeyun Gong and
                  Lionel M. Ni and
                  Heung{-}Yeung Shum and
                  Jian Guo},
  title        = {Think-on-Graph: Deep and Responsible Reasoning of Large Language Model
                  on Knowledge Graph},
  booktitle    = {{ICLR}},
  publisher    = {OpenReview.net},
  year         = {2024},
  url          = {https://openreview.net/forum?id=nnVO1PvbTv}
}

@article{bhatta2024use,
  title={Use of IoT for structural health monitoring of civil engineering structures: a state-of-the-art review},
  author={Bhatta, Sanjeev and Dang, Ji},
  journal={Urban Lifeline},
  volume={2},
  number={1},
  pages={17},
  year={2024},
  publisher={Springer}
}

@article{janowicz2019sosa,
  author       = {Krzysztof Janowicz and
                  Armin Haller and
                  Simon J. D. Cox and
                  Danh Le Phuoc and
                  Maxime Lefran{\c{c}}ois},
  title        = {{SOSA:} {A} lightweight ontology for sensors, observations, samples,
                  and actuators},
  journal      = {J. Web Semant.},
  volume       = {56},
  pages        = {1--10},
  year         = {2019},
  url          = {https://doi.org/10.1016/j.websem.2018.06.003},
  doi          = {10.1016/J.WEBSEM.2018.06.003}
}

@article{haller2018modular,
  author       = {Armin Haller and
                  Krzysztof Janowicz and
                  Simon J. D. Cox and
                  Maxime Lefran{\c{c}}ois and
                  Kerry Taylor and
                  Danh Le Phuoc and
                  Joshua Lieberman and
                  Ra{\'{u}}l Garc{\'{\i}}a{-}Castro and
                  Rob Atkinson and
                  Claus Stadler},
  title        = {The modular {SSN} ontology: {A} joint {W3C} and {OGC} standard specifying
                  the semantics of sensors, observations, sampling, and actuation},
  journal      = {Semantic Web},
  volume       = {10},
  number       = {1},
  pages        = {9--32},
  year         = {2019},
  url          = {https://doi.org/10.3233/SW-180320},
  doi          = {10.3233/SW-180320}
}

@inproceedings{xiao2020virtual,
  author       = {Guohui Xiao and
                  Davide Lanti and
                  Roman Kontchakov and
                  Sarah Komla{-}Ebri and
                  Elem G{\"{u}}zel Kalayci and
                  Linfang Ding and
                  Julien Corman and
                  Benjamin Cogrel and
                  Diego Calvanese and
                  Elena Botoeva},
  title        = {The Virtual Knowledge Graph System Ontop},
  booktitle    = {{ISWC} {(2)}},
  series       = {Lecture Notes in Computer Science},
  volume       = {12507},
  pages        = {259--277},
  publisher    = {Springer},
  year         = {2020},
  url          = {https://doi.org/10.1007/978-3-030-62466-8\_17},
  doi          = {10.1007/978-3-030-62466-8\_17}
}

@article{doerr2003cidoc,
  author       = {Martin Doerr},
  title        = {The {CIDOC} Conceptual Reference Module: An Ontological Approach to
                  Semantic Interoperability of Metadata},
  journal      = {{AI} Mag.},
  volume       = {24},
  number       = {3},
  pages        = {75--92},
  year         = {2003},
  url          = {https://doi.org/10.1609/aimag.v24i3.1720},
  doi          = {10.1609/AIMAG.V24I3.1720}
}

@inproceedings{wang2022self,
  author       = {Xuezhi Wang and
                  Jason Wei and
                  Dale Schuurmans and
                  Quoc V. Le and
                  Ed H. Chi and
                  Sharan Narang and
                  Aakanksha Chowdhery and
                  Denny Zhou},
  title        = {Self-Consistency Improves Chain of Thought Reasoning in Language Models},
  booktitle    = {{ICLR}},
  publisher    = {OpenReview.net},
  year         = {2023},
  url          = {https://openreview.net/forum?id=1PL1NIMMrw}
}

@inproceedings{pavlidis2025panoptes,
  title={PANOPTES: A Digital Twin Ontology for Cultural Asset Management},
  author={Pavlidis, George and Sevetlidis, Vasileios and Arampatzakis, Vasileios},
  booktitle={2025 IEEE International Conference on Cyber Humanities (IEEE-CH)},
  pages={1--6},
  year={2025},
  organization={IEEE}
}

@article{rossi2023structural,
  title={Structural health monitoring and management of cultural heritage structures: a state-of-the-art review},
  author={Rossi, Michela and Bournas, Dionysios},
  journal={Applied Sciences},
  volume={13},
  number={11},
  pages={6450},
  year={2023},
  publisher={MDPI}
}

@article{vasic2025knowledge,
  author       = {Iva Vasic and
                  Hans{-}Georg Fill and
                  Ramona Quattrini and
                  Roberto Pierdicca},
  title        = {Knowledge Graphs vs. Large Language Models: Competitors or Partners
                  in Supporting Virtual Museums},
  journal      = {{ACM} Journal on Computing and Cultural Heritage},
  volume       = {18},
  number       = {4},
  pages        = {61:1--61:28},
  year         = {2025},
  url          = {https://doi.org/10.1145/3756016},
  doi          = {10.1145/3756016}
}

@article{niccolucci2024digital,
  author       = {Franco Niccolucci and
                  Achille Felicetti},
  title        = {Digital Twin Sensors in Cultural Heritage Ontology Applications},
  journal      = {Sensors},
  volume       = {24},
  number       = {12},
  pages        = {3978},
  year         = {2024},
  url          = {https://doi.org/10.3390/s24123978},
  doi          = {10.3390/S24123978}
}

@article{xu2024nanjing,
  title={Nanjing Yunjin intelligent question-answering system based on knowledge graphs and retrieval augmented generation technology},
  author={Xu, Liang and Lu, Lu and Liu, Minglu and Song, Chengxuan and Wu, Lizhen},
  journal={Heritage Science},
  volume={12},
  number={1},
  pages={118},
  year={2024},
  publisher={Nature Publishing Group}
}

@article{suissa2023around,
  author       = {Omri Suissa and
                  Maayan Zhitomirsky{-}Geffet and
                  Avshalom Elmalech},
  title        = {Around the {GLOBE:} Numerical Aggregation Question-answering on Heterogeneous
                  Genealogical Knowledge Graphs with Deep Neural Networks},
  journal      = {{ACM} Journal on Computing and Cultural Heritage},
  volume       = {16},
  number       = {3},
  pages        = {51:1--51:24},
  year         = {2023},
  url          = {https://doi.org/10.1145/3586081},
  doi          = {10.1145/3586081}
}

@inproceedings{wang2024chain,
  author       = {Zilong Wang and
                  Hao Zhang and
                  Chun{-}Liang Li and
                  Julian Martin Eisenschlos and
                  Vincent Perot and
                  Zifeng Wang and
                  Lesly Miculicich and
                  Yasuhisa Fujii and
                  Jingbo Shang and
                  Chen{-}Yu Lee and
                  Tomas Pfister},
  title        = {Chain-of-Table: Evolving Tables in the Reasoning Chain for Table Understanding},
  booktitle    = {{ICLR}},
  publisher    = {OpenReview.net},
  year         = {2024},
  url          = {https://openreview.net/forum?id=4L0xnS4GQM}
}

@inproceedings{ye2024flask,
  author       = {Seonghyeon Ye and
                  Doyoung Kim and
                  Sungdong Kim and
                  Hyeonbin Hwang and
                  Seungone Kim and
                  Yongrae Jo and
                  James Thorne and
                  Juho Kim and
                  Minjoon Seo},
  title        = {{FLASK:} Fine-grained Language Model Evaluation based on Alignment
                  Skill Sets},
  booktitle    = {{ICLR}},
  publisher    = {OpenReview.net},
  year         = {2024},
  url          = {https://openreview.net/forum?id=CYmF38ysDa}
}

@inproceedings{zheng2023judging,
  author       = {Lianmin Zheng and
                  Wei{-}Lin Chiang and
                  Ying Sheng and
                  Siyuan Zhuang and
                  Zhanghao Wu and
                  Yonghao Zhuang and
                  Zi Lin and
                  Zhuohan Li and
                  Dacheng Li and
                  Eric P. Xing and
                  Hao Zhang and
                  Joseph E. Gonzalez and
                  Ion Stoica},
  title        = {Judging LLM-as-a-Judge with MT-Bench and Chatbot Arena},
  booktitle    = {NeurIPS},
  year         = {2023},
  url          = {http://papers.nips.cc/paper\_files/paper/2023/hash/91f18a1287b398d378ef22505bf41832-Abstract-Datasets\_and\_Benchmarks.html}
}

@inproceedings{chen2023multi,
  author       = {Ziyang Chen and
                  Jinzhi Liao and
                  Xiang Zhao},
  title        = {Multi-granularity Temporal Question Answering over Knowledge Graphs},
  booktitle    = {{ACL} {(1)}},
  pages        = {11378--11392},
  publisher    = {Association for Computational Linguistics},
  year         = {2023},
  url          = {https://doi.org/10.18653/v1/2023.acl-long.637},
  doi          = {10.18653/V1/2023.ACL-LONG.637}
}

@article{harikumar2026plancompiler,
  author       = {Pranav Harikumar},
  title        = {PlanCompiler: {A} Deterministic Compilation Architecture for Structured
                  Multi-Step {LLM} Pipelines},
  journal      = {CoRR},
  volume       = {abs/2604.13092},
  year         = {2026},
  url          = {https://doi.org/10.48550/arXiv.2604.13092},
  doi          = {10.48550/ARXIV.2604.13092}
}

@article{zhang2026saga,
  author       = {Yiming Zhang and
                  Koji Tsuda},
  title        = {{SAGA:} Schema-Aware Grounding for Agentic Text-to-SPARQL Generation},
  journal      = {CoRR},
  volume       = {abs/2607.14494},
  year         = {2026},
  url          = {https://doi.org/10.48550/arXiv.2607.14494},
  doi          = {10.48550/ARXIV.2607.14494}
}

@inproceedings{schwarzinger2025sigsparql,
  author       = {Tobias Schwarzinger and
                  Gernot Steindl and
                  Thomas Fr{\"{u}}hwirth and
                  Thomas Preindl and
                  Konrad Diwold and
                  Katrin Ehrenm{\"{u}}ller and
                  Fajar J. Ekaputra},
  title        = {SigSPARQL: Signals as a First-Class Citizen when Querying Knowledge
                  Graphs},
  booktitle    = {SEMANTiCS},
  series       = {Studies on the Semantic Web},
  volume       = {62},
  pages        = {159--175},
  publisher    = {{IOS} Press},
  year         = {2025},
  url          = {https://doi.org/10.3233/SSW250018},
  doi          = {10.3233/SSW250018}
}

@article{khan2025data,
  title={From Data Center IoT Telemetry to Data Analytics Chatbots--Virtual Knowledge Graph is All You Need},
  author={Khan, Junaid Ahmed and Cavagna, Hiari Pizzini and Proia, Andrea and Bartolini, Andrea},
  journal={arXiv preprint arXiv:2506.22267},
  year={2025}
}
